%% file: root.tex
\documentclass[letterpaper, 10 pt, conference]{ieeeconf}
\IEEEoverridecommandlockouts
\usepackage{amsmath,amssymb,amsfonts}
\usepackage{graphicx}
\usepackage{textcomp}
\usepackage{xcolor}
\usepackage{booktabs}
\usepackage{algorithm}
\usepackage{cleveref}
\usepackage{pgfplots}
\usepgfplotslibrary{groupplots}
\usepackage[noend]{algpseudocode}
\usepackage{wrapfig}

\usepackage{carshapes}

\usetikzlibrary{positioning,arrows}
\usepgfplotslibrary{fillbetween}

\pgfplotsset{compat=newest}
\pgfplotsset{every axis legend/.append style={legend cell align=left}}

\pgfplotsset{every axis/.append style={
                    title style={font=\small},
                    tick label style={font=\footnotesize}  
                    }}
\pgfplotsset{every axis label/.style={font=\small}}                    
\pgfplotsset{
legend image code/.code={
\draw[mark repeat=2,mark phase=2]
plot coordinates {
(0cm,0cm)
(0.15cm,0cm)        
(0.3cm,0cm)         
};%
}
}

\pgfplotsset{
xlabel near ticks/.style={
every axis x label/.style={
    at={(ticklabel cs:0.5)},anchor=near ticklabel
     }
  },
ylabel near ticks/.style={
every axis y label/.style={
    at={(ticklabel cs:0.5)},rotate=90,anchor=near ticklabel}
     }
  }
  
\definecolor{pastelMagenta}{HTML}{FF48CF}
\definecolor{pastelPurple}{HTML}{8770FE}
\definecolor{pastelBlue}{HTML}{1BA1EA}
\definecolor{pastelSeaGreen}{HTML}{14B57F}
\definecolor{pastelGreen}{HTML}{3EAA0D}
\definecolor{pastelOrange}{HTML}{C38D09}
\definecolor{pastelRed}{HTML}{F5615C}

\input{pastelcolormap.tex}
\input{pastelcolorlist.tex}

\usepackage[keeplastbox]{flushend}

\usepackage[backend=bibtex,style=ieee,mincitenames=1,maxcitenames=2,maxbibnames=20]{biblatex}
\def\BibTeX{{\rm B\kern-.05em{\sc i\kern-.025em b}\kern-.08em
    T\kern-.1667em\lower.7ex\hbox{E}\kern-.125emX}}
\begin{document}

\title{\LARGE \bf
Preference-based Learning of Reward Function Features
}

\author{Sydney M. Katz$^{*}$, Amir Maleki$^{*}$, Erdem B{\i}y{\i}k, and Mykel J. Kochenderfer
\thanks{*Denotes equal contribution}
\thanks{\textit{Stanford University}, Stanford, CA 94305, \{smkatz, amir.maleki, ebiyik, mykel\}@stanford.edu}%
}

\maketitle

\begin{abstract}
Preference-based learning of reward functions, where the reward function is learned using comparison data, has been well studied for complex robotic tasks such as autonomous driving. Existing algorithms have focused on learning reward functions that are linear in a set of trajectory features. The features are typically hand-coded, and preference-based learning is used to determine a particular user's relative weighting for each feature. Designing a representative set of features to encode reward is challenging and can result in inaccurate models that fail to model the users' preferences or perform the task properly. In this paper, we present a method to learn both the relative weighting among features as well as additional features that help encode a user's reward function. The additional features are modeled as a neural network that is trained on the data from pairwise comparison queries. We apply our methods to a driving scenario used in previous work and compare the predictive power of our method to that of only hand-coded features. We perform additional analysis to interpret the learned features and examine the optimal trajectories. Our results show that adding an additional learned feature to the reward model enhances both its predictive power and expressiveness, producing unique results for each user. 
\end{abstract}

\section{Introduction} \label{sec:intro}
If designed properly, reward functions provide a means for humans to convey desired behavior to a robot. However, creating a reward function that accurately encodes human intent can be difficult in complicated, high-dimensional problem settings. Preference-based learning has been proposed as a way to address this challenge.  By querying the human user with a set of demonstrated trajectories and asking which demonstration they prefer, the robot learns the reward function it should optimize.
Its motivation is similar to that of inverse reinforcement learning (IRL) in that it allows humans to train a policy without explicitly specifying the values of the reward function parameters \cite{ng2000algorithms}. For this reason, preference-based learning has been successfully applied to challenging problems in robotics \cite{wirth2017survey,wilson2012bayesian,christiano2017deep,akrour2011preference,cakmak2011human,ibarz2018reward,brown2019deep,wilde2019bayesian,tucker2020preference,biyik2020active,biyik2019asking,palan2019learning,wilde2020active,sadigh2017active,biyik2018batch}.

While some preference-based learning algorithms focus on learning a policy directly \cite{wilson2012bayesian, christiano2017deep}, others focus on learning a reward or utility function \cite{sadigh2017active, biyik2019asking, biyik2018batch, palan2019learning, katz2019learning, biyik2020active, alg4opt, lepird2015bayesian, tucker2020preference,wilde2020active}, which typically involves a lower-dimensional parameter space to make results easier to interpret. Current state-of-the-art algorithms provide methods to learn reward functions that are linear in a set of features \cite{sadigh2017active, biyik2018batch, katz2019learning, palan2019learning, shah2019preferences, bajcsy2017learning, biyik2019asking,wilde2020active}. Features are hand-coded functions of a robot trajectory that remain constant throughout the learning process. Responses to preference queries are used to learn weights that indicate the relative importance of each feature. For example, in a driving scenario, preference-based learning has been used to determine the relative importance of features such as collision avoidance and keeping speed by obtaining pairwise preferences over driving trajectories \cite{sadigh2017active}.

While there have been studies to relax the linearity assumption by modeling reward as a mixture of linear functions conditioned on latent states of the user \cite{basu2019active} or as a Gaussian process \cite{biyik2020active}, the features were still hand-designed. Because reward functions are based entirely on hand-coded features, the features must be expressive enough to model human intent and can be difficult to design appropriately \cite{sadigh2017active, biyik2018batch, katz2019learning}. Feature selection presents three main challenges: selecting representative features, selecting expressive features, and selecting the proper functional form. Features must be representative enough to accurately model the reward function for a particular user. \citeauthor{sadigh2017active} \cite{sadigh2017active} found that applying preference-based learning to understand driving behavior resulted in similar reward functions among study participants. In order to distinguish between specific users' personal driving preferences, a more expressive set of features may be required \cite{sadigh2017active}. Finally, applying preference-based learning to a variety of tasks requires complex, hand-tuned, nonlinear feature functions, and designing them is challenging \cite{biyik2019asking}.

In this work, we address some of these challenges
by providing a framework to not only learn the weights of a linear reward function but also to learn additional nonlinear features. We represent features as a neural network that is trained on a user's responses to queries. We apply the learning framework to the driving problem originally presented by \citeauthor{sadigh2017active} \cite{sadigh2017active} and compare the results using the learned features to those using only hand-coded features. Not only do we evaluate the model's predictive power, but we also visualize and interpret the features learned by the neural network. Our results demonstrate that an additional nonlinear feature improves the predictive power of the framework and enables it to better personalize users' preferences.


\section{Approach} \label{sec:approach}
Our approach to feature learning requires two phases. The first involves obtaining expert preferences using an active querying method. The second is to use the preferences to learn features. Let $\phi(\mathbf{x}^t)$ be a function that maps the state $\mathbf{x}$ of a robotic system at time $t$ to a set of features. The reward function takes the form
\begin{equation}
    r(\mathbf{x}^t) = \mathbf{w}^\top\phi(\mathbf{x}^t)
\end{equation}
where $\mathbf{w}$ is a vector that specifies the relative weighting of each feature. Defining a trajectory $\tau$ to consist of an initial state and a set of $k$ subsequent states, we let $\Phi(\tau) = \sum_{t=0}^k \phi(\mathbf{x}^t)$. The reward over a particular trajectory is then
\begin{equation}
    R(\tau) = \mathbf{w}^\top\Phi(\tau)
\end{equation}
In order to preserve the interpretability of the final model, we select a mixed feature model in which the feature function contains both hand-coded and neural network features. Thus,
\begin{equation}
   r(\mathbf{x}^t) = [\mathbf{w}_{hc}, \mathbf{w}_{nn}]^\top [\phi_{hc}(\mathbf{x}^t), \phi_{nn}(\mathbf{x}^t)]  
\end{equation}
where $\phi_{hc}(\mathbf{x}^t)$ represents the hand-coded feature function with corresponding weight vector $\mathbf{w}_{hc}$, and $\phi_{nn}(\mathbf{x}^t)$ represents the neural network feature function with corresponding weight vector $\mathbf{w}_{nn}$. 

The methods outlined in our approach discuss techniques to learn $\mathbf{w}_{hc}$, $\mathbf{w}_{nn}$, and the function $\phi_{nn}(\mathbf{x}^t)$. We note that requiring hand-coded features still presents a challenge in feature design. However, learning additional features using a neural network can capture important trajectory qualities that the feature designer may have missed.

\subsection{Preference Model}
Preference-based learning is an iterative process between querying the user for their preference and updating our model.
We keep a distribution $p(\mathbf{w}$) over possible values of $\mathbf{w}$ and perform Bayesian updates to it as we obtain preferences. Let the $n$th pairwise comparison query contain the trajectories $\tau^{(n)}_a$ and $\tau^{(n)}_b$. We define $I_n$ as the response to the $n$th query, where
\begin{equation}
    I_n = \begin{cases}
    +1, & \tau^{(n)}_a \succ \tau^{(n)}_b \\
    -1, & \tau^{(n)}_a \prec \tau^{(n)}_b
    \end{cases}
\end{equation}
with $\tau_a \succ \tau_b$ representing the user's preference of $\tau_a$ to $\tau_b$. The Bayesian update can be written as follows:
\begin{equation} \label{eq:bayesian_update}
    p(\mathbf{w} \mid I_n) \propto p(I_n \mid \mathbf{w}) p(\mathbf{w})
\end{equation}
where $p(\mathbf{w})$ encodes the current distribution over $\mathbf{w}$ that takes into account responses $I_{1:n-1}$. Before any preferences have been obtained, we assume a uniform prior over the search space of possible values.

In order to perform this update, we must specify a likelihood model for $p(I_n \mid \mathbf{w})$ keeping in mind that we expect occasional errors from the user. As in \citeauthor{sadigh2017active} \cite{sadigh2017active}, we use a sigmoid likelihood function:
\begin{equation} \label{eq:siglike}
    p(I_n \mid \mathbf{w}) = \frac{1}{1 + \exp[-I_n(R(\tau_a^{(n)}) - R(\tau_b^{(n)}))]}
\end{equation}
Samples from the posterior $p(\mathbf{w} \mid I_n)$ can be generated using Markov Chain Monte Carlo (MCMC) methods. In particular, we use the adaptive Metropolis algorithm to efficiently generate samples at each iteration \cite{haario2001adaptive}.
 
\subsection{Active Querying} \label{sec:queries}
Active querying methods are desirable because they decrease the number of required queries by asking questions based on current model uncertainty inferred from prior user preferences. Methods for active querying have been proposed in the literature \cite{sadigh2017active, biyik2018batch, biyik2019asking, katz2019learning, wilde2020active}. These methods typically rely on solving an optimization problem over a continuous action space or control input \cite{sadigh2017active, biyik2018batch, biyik2019asking}. One objective is to select the set of control inputs that maximizes the volume removed from the current distribution over model parameters \cite{sadigh2017active, biyik2018batch}. In later work, better performance was achieved using an objective that seeks to maximize the information gained from each query \cite{biyik2019asking}.

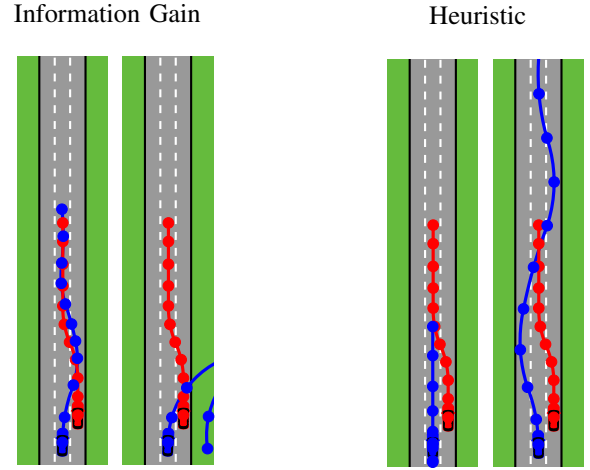
\begin{figure}[htb]
    \centering
    \input{ig_hs_traj.tex}
    \caption{\label{fig:ig_trajs} Example queries with information gain objective (left) and heuristic method (right). Marks are placed at equal time intervals on all trajectories. The car in the right panel of the information gain query leaves the road and backs up to the starting latitude.}
\end{figure}

Although these objectives generate queries that are maximally informative, they do not provide a direct incentive to generate query trajectories that are realistic. \Cref{fig:ig_trajs} shows an example of two possible queries that can be shown to a user for a driving scenario. The user is presented with two options and is asked to select their preferred trajectory for the blue vehicle.

The query on the left of \cref{fig:ig_trajs} is generated using the information gain objective \cite{biyik2019asking}. The user is asked to select between a scenario in which the blue vehicle repeatedly crashes into the other vehicle and a scenario in which the blue vehicle drives off the road and begins to drive backward. While these trajectories are informative with respect to the collision avoidance and staying within the lanes on the road, neither trajectory is realistic. Because we want to learn features that allow us to express specific user intentions, it is especially important to show users realistic trajectories during the active learning process. For this reason, we adopt an approach that shows users trajectories that have been optimized for reward functions defined by different vectors $\mathbf{w}$ \cite{katz2019learning}. By maximizing a reward function for each trajectory, we ensure that the trajectories will have realistic features. A query generated using this approach is shown in the right half of \cref{fig:ig_trajs}.

To choose these $\mathbf{w}$  vectors from the samples from $p(\mathbf{w})$, we formulate a multiobjective optimization problem similar to prior work \cite{wilson2012bayesian, katz2019learning}. Let $M$ be the number of MCMC samples generated after each Bayesian update, and let $\mathbf{w}_i$ be the $i$th sample in this set. The optimization problem can be written as
\begin{equation}
    \underset{i,j \text{ s.t. } i \neq j}{\text{maximize }} p(\mathbf{w}_i)p(\mathbf{w}_j) + \mu \| \mathbf{w}_i - \mathbf{w}_j \|_2
\end{equation}
where $\mu \geq 0$ controls the balance between the objectives. The first term incentivizes selecting weights that are likely based on the current estimate of $p(\mathbf{w})$, and the second term ensures we select samples that are different enough to produce a distinguishable query. To calculate the first term, we use unnormalized posteriors obtained by \cref{eq:bayesian_update}.

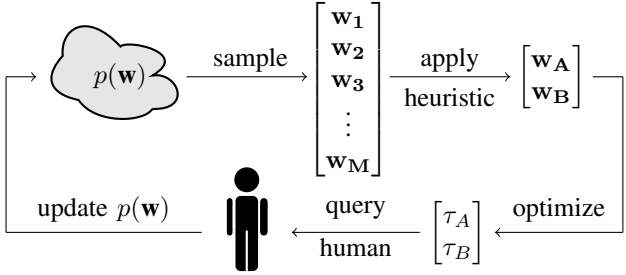
\begin{figure}[t]
    \centering
    \input{active_query_process.tex}
    \caption{Overview of the iterative learning process. \label{fig:query_gen_process}}
\end{figure}

\Cref{fig:query_gen_process} outlines the overall active querying process. First, samples are drawn from $p(\mathbf{w})$ using MCMC. Next, we apply our multiobjective optimization heuristic to select two of these samples to use for the next query. Query trajectories are generated by maximizing the reward function of each sample and subsequently shown to the user. The optimization is nonconvex and is solved using the quasi-Newton method L-BFGS \cite{alg4opt}. We solve the optimization problem 10 times starting from random initial points and show the query with the best objective value. After we obtain the user's preference, we update the posterior $p(\mathbf{w})$ and the process resumes with sampling.

\subsection{Feature Learning}
We seek to learn neural network features to augment our feature function $\Phi(\tau)$ so that there exists a weight vector $\mathbf{w}$ such that the linear function $\mathbf{w}^\top \Phi(\tau)$ is a good predictor of reward. More specifically, we want to predict higher reward for the trajectory that the user selected in each query. Note that in the mixed feature setting, $\phi(\mathbf{x}^t)$ is influenced not only by $\phi_{nn}(\mathbf{x}^t)$ but also by $\phi_{hc}(\mathbf{x}^t)$, and we must take this into account in our training.

\subsubsection{Network Structure}
Our learned feature function $\phi_{nn}(\mathbf{x}^t)$ is represented by a neural network in which the input is a function of the state of the system. For a simple system, this function can be the identity mapping (i.e. the input to the network is the state of the system). The final layer of the network contains one neuron per feature and represents the learned features. Because the hand-coded features are normalized to have a magnitude less than or equal to 1 when averaged over the trajectory, we select hyperbolic tangent as the activation function for the output layer. This activation ensures that features all have similar magnitude.

\subsubsection{Loss Function}
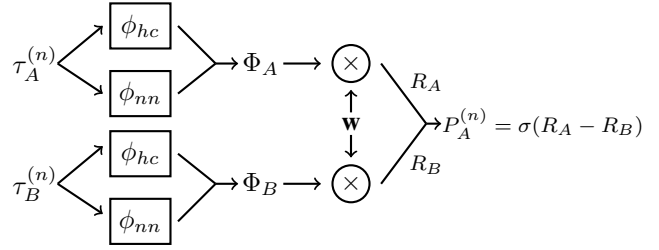
\begin{figure}[t]
    \vspace{0.3cm}
    \centering
    \input{compute_prob.tex}
    \caption{Estimating the probability of selecting trajectory A for a particular query based on the current feature functions and weights. The $\sigma$ in the last step represents a sigmoid function. \label{fig:comp_prob}}
\end{figure}

The feature learning problem can be thought of as a classification problem in that we would like to find features that allow us to make binary predictions on queries regarding whether or not a user prefers trajectory A to trajectory B. Framing the problem in this way allows us to use a cross entropy loss function that takes in the probability of the user selecting $\tau_A$ given the current reward function model:
\begin{equation}
    \text{loss} = \frac{1}{N} \sum_{n=1}^N \left[ z_n \log(P_A^{(n)}) - (1 - z_n) \log(1 - P_A^{(n)}) \right]
\end{equation}
where $z_n = I_n$ if $I_n = 1$, $z_n = 0$ if $I_n = -1$, $P_A^{(n)}$ is the probability of the user selecting $\tau_A^{(n)}$ over $\tau_B^{(n)}$, and $N$ is the number of preference queries in the training set. \Cref{eq:siglike} can be used to calculate $P_A^{(n)}$, and \cref{fig:comp_prob} provides a visual representation of this process.
We allow our model to train both the neural network parameters and the linear reward weights $\mathbf{w}$.

\section{Experimental Setup} \label{sec:exp}
\subsection{Problem Domain}
We tested our approach on the driving simulation\footnote{Source is at \url{github.com/sisl/FeatureLearningPrefs}.} that has been used in previous preference-based learning works \cite{sadigh2017active, biyik2018batch, biyik2019asking, wilde2020active}. The right panels in \cref{fig:ig_trajs} show an example of a preference query that would be shown to a user. The red car, which represents the human-driven vehicle, starts in the rightmost lane and switches to the middle lane. The human-driven vehicle follows the same trajectory in every scenario. The user is asked to select the trajectory they would prefer the blue car, or robot-driven vehicle, to follow. The state is defined as $[x_r, y_r, \theta_r, v_r, x_h, y_h, \theta_h, v_h]$ where $x$ is the vehicle's latitudinal position on the road, $y$ is the vehicle's longitudinal position, $\theta$ is the vehicle's heading ($\theta = 90^\circ$ corresponds to driving straight along the road), and $v$ is the vehicle's speed. A subscript $r$ denotes a state corresponding to the robot car, and a subscript $h$ denotes a state associated with the human car.

\subsubsection{Hand-coded Features}
The hand-coded features are simlar to those used in \citeauthor{biyik2019asking} \cite{biyik2018batch} 
In order to narrow our search space of vectors $\mathbf{w}$, we design our features such that higher values are preferred and restrict the elements of $\mathbf{w}$ to be nonnegative. The features are summarized in \cref{tab:hc_feats}. 
\begin{table}[htb]
    \centering
    \caption{Hand-coded Features \label{tab:hc_feats}}
    \begin{tabular}{@{}ll@{}}
         \toprule
         \textbf{Description} & \textbf{Expression} \\
         \midrule
         \addlinespace[0.2em]
         staying in lane & $\exp \left[\min((x_r - 0.17)^2, x_r^2, (x_r + 0.17)^2) \right]$ \\ 
         \addlinespace[0.2em]
         keeping speed & $-(v_r - 1)^2$ \\
         \addlinespace[0.2em]
         heading & $\sin(\theta_r)$ \\
         \addlinespace[0.2em]
         collision avoidance & $-\exp\left[-(7(x_r - x_h)^2 + 3(y_r - y_h)^2) \right]$ \\
         \bottomrule
    \end{tabular}
\end{table}

In this particular scenario, lane centers are located at $x = -0.17$, $x = 0$, and $x = 0.17$, so the first feature represents the minimum distance to a lane center. The collision avoidance feature was designed for a vehicle with an aspect ratio of 7/3. These features are expressive enough to produce a reasonable optimal trajectory given a weight vector $\mathbf{w}$; however, we hypothesize that this set of features alone is not expressive enough to fully describe a driver's reward function and differentiate driving styles \cite{sadigh2017active, biyik2019asking}.

\subsubsection{Neural Network Structure and Inputs}
For baseline testing, the input to our neural network is a function of the robot and human cars' states. Specifically, we input $x_r$, $\theta_r$, and $v_r$ from the robot state. However, while $x_r$ provides an indication of whether or not the car is staying in its lane, $y_r$ does not provide relevant information on its own. Instead, we to provide the network with the distance between the robot car and human car defined as
\begin{equation}
    d = \sqrt{(x_r - x_h)^2 + (y_r - y_h)^2}
\end{equation}
Thus, our neural network input is the vector $[x_r, d, \theta_r, v_r]$. 
After initial testing, we experimented with adding an extra input to the network as discussed in \cref{sec:res}. 
In this experiment, we learn one extra feature to augment the hand-coded features, so the final layer of the network consists of a single neuron. The network is a fully connected feedforward network with one hidden layer containing 100 neurons with Rectified Linear Unit (ReLU) activation functions.
We tested the effect of using more than one extra feature. However, we found the learned features to be redundant without a significant performance improvement. A single neural network feature was complex enough to capture many aspects of driving (see \cref{sec:res}).

\subsection{Training Details}
While the linear reward weight for the neural network feature $\mathbf{w}_{nn}$ and neural network parameters were randomly initialized, the linear reward weights for the hand-coded features $\mathbf{w}_{hc}$ were initialized to the estimates obtained during the active query process. For each training epoch, the gradient of the loss function is computed with respect to the neural network parameters, and the neural network parameters are updated using the NADAM algorithm with a learning rate of 0.001 and $\beta$ parameters of (0.9, 0.999) \cite{dozat2016incorporating}. 

As the neural network parameters are updated and its output feature value changes relative to the hand-coded feature values, the linear reward weights may no longer be optimal. For this reason, we update the linear reward  during training as well; however, because the linear reward weights have already been learned from the active querying process, we found that we could improve learning by only updating them every 20 epochs via gradient descent.
We divide the data into training, validation, and test sets and repeat the training process over 40 trials, resetting the linear reward weights between trials. We select the 5 neural networks out of the 40 trials that perform best at prediction on the validation set for our evaluations on the test set.

\subsection{Dependent Measures}
We assess the \emph{predictive power} of the models by evaluating the ratio of user comparisons (in the test set) that are correctly predicted by the learned reward function. We also qualitatively evaluate the model's ability to generate safe,  customized optimal trajectories that reflect the different driving preferences of the users.

\subsection{Hypotheses}
We test the following hypotheses with our experiments. Using the learned neural network features in addition to the hand-coded features (\textbf{H1}) \textit{improves the predictive power of the reward model} and (\textbf{H2}) \textit{enables better customized optimal driving trajectories}. 

\subsection{User Study Procedure}
To test the hypotheses, we conducted a user study with 15 participants, all of whom have a valid driver's license. Each participant was shown 100 actively generated queries using the method described in \cref{sec:queries} and the hand-coded features. Because the queries are actively generated based on user responses, each participant was shown a different set of trajectories during this part of the study. For the next part of the study, we created a standardized set of 75 queries as a test set.
The queries in the test set were generated by sampling random pairs of values for $\mathbf{w}_{hc}$ and solving for the locally optimal trajectories.

For neural network training, we use the first 70 trajectory pairs presented to the user as the training data with the preferences obtained from the user as the training labels. We use the remaining 30 as the validation set. The 75 standardized trajectory pairs from the second part of the study and corresponding user responses serve as the test data and labels. We compared our method, which involves both the hand-coded and the learned features, with the method in prior works that uses only hand-coded features.

\section{Results} \label{sec:res}
We analyzed the neural network features from the user study for both their predictive power and interpretability. 

\subsection{Predictive Power}
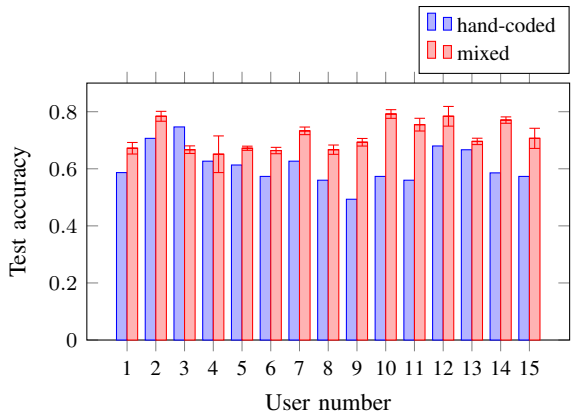
\begin{figure}[htb]
    \centering
    \input{accuracy_chart.tex}
    \caption{Test accuracy of hand-coded and mixed feature sets for each user. While obtaining the hand-coded test accuracy (blue bars) is a deterministic process, the red bar heights represent the mean of the test accuracy of the neural networks that had the top 5 validation accuracy values. Error bars show standard deviation among the 5 neural network trials. \label{fig:accs}}
\end{figure}

We analyzed the prediction accuracy of both the hand-coded and mixed feature sets for each user on the standardized test set. \Cref{fig:accs} shows the results for each user. For all users except user 3, the mixed feature set shows an improvement over the hand-coded only accuracy. Improvements range from 2\% to 21\%. This result strongly supports \textbf{H1}. Users with small improvements tended to have high test accuracy using the hand-coded only features. A high test accuracy with the hand-coded features indicates that the hand-coded features alone provide an adequate encoding of the user's reward function, and further learning may not be necessary. For instance, user 3 had the highest hand-coded test accuracy of all users in the study, which may explain the inability to learn a feature that improves the predictive power of their reward function. On the other hand, the mixed feature set showed significant increases in performance for users whose preferences could not be fully described by the hand-coded features.


\subsection{Feature Interpretation}
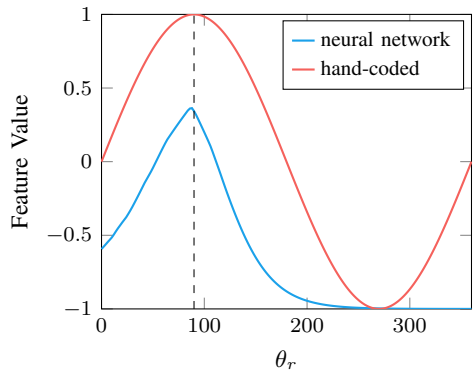
\begin{figure}[htb]
    \centering
    \input{heading_both.tex}
    \caption{Variation in neural network feature value and hand-coded heading feature value with heading for user 1. The blue line shows the output of the neural network when $x_r=0.0$, $d=0.5$, and $v_r=0.8$ and $\theta_r$ is varied from 0 to 360 degrees. The red line represents $\sin \theta_r$, the hand-coded feature for heading. The dashed line shows a heading of $90^\circ$, corresponding to driving straight down the road. \label{fig:theta_user1}}
\end{figure}

Because the input to the neural network is four dimensional, we can only visualize the feature values for slices of the input space. \Cref{fig:theta_user1} shows the change in the neural network feature value for changes in heading for user 1 and compares it with the hand-coded feature for maintaining heading. The vehicle is located in the center lane at a distance $d=0.5$ from the other vehicle and is traveling at 80\% of maximum speed.

It is clear that the neural network feature for user 1 is influenced by vehicle heading. The feature value has a sharp peak around $90^\circ$, which corresponds to driving straight down the road. Because reward weights are positive, a higher neural network feature value indicates a greater positive contribution to the total reward. Therefore, the neural network feature will contribute to a higher reward for trajectories with vehicle headings near $90^\circ$. While the hand-coded heading feature penalizes facing backward (heading of 270 degrees) more than it penalizes facing left and right (heading of 0 or 180 degrees), the neural network feature does not show the same trend. Instead, it has a sharp peak around $90^\circ$ but then quickly decreases to the minimum value.



\begin{figure}[htb]
    \vspace{0.2cm}
    \centering
    \input{feature_viz_2d.tex}
    \caption{Heat map of neural network feature value for various locations of the human-driven vehicle (represented by the red car) for user 1. The speed is held constant at $v_r=1$, and heading is held constant at $\theta_r=90^\circ$. Brighter colors indicate higher feature values. \label{fig:viz_2d}}
\end{figure}
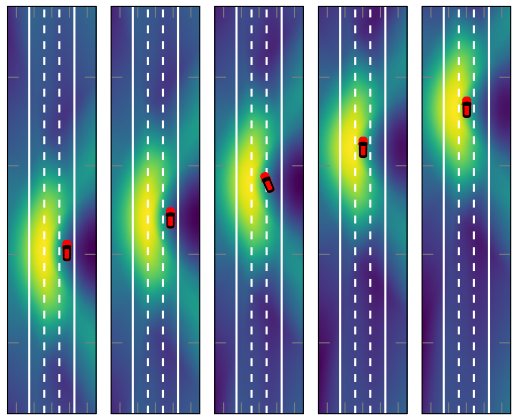

\Cref{fig:viz_2d} shows the variation in the neural network feature value for user 1 for various locations of the human-driven vehicle on the road. Speed is held constant at the maximum speed, and heading is held constant at $90^\circ$.
This visualization demonstrates that the neural network feature has a form of collision avoidance built into it. The feature value decreases in close proximity to the other vehicle on the road. Additionally, the value drops off outside the lanes of the road. The value is highest in the region of the road just to the left of the other vehicle. This trend can be interpreted as a preference to maintain speed and keep up with traffic.

\begin{figure}[htb]
    \centering
    \input{feature_viz_2d_time_gap.tex}
    \caption{Heat map of neural network feature value with augmented input space for various locations of the other vehicle (represented by the red car) for user 1. The speed is held constant at $v_r=1$, and heading is held constant at $\theta_r=90^\circ$. Brighter colors indicate higher feature values. \label{fig:viz_2d_time_gap}}
\end{figure}
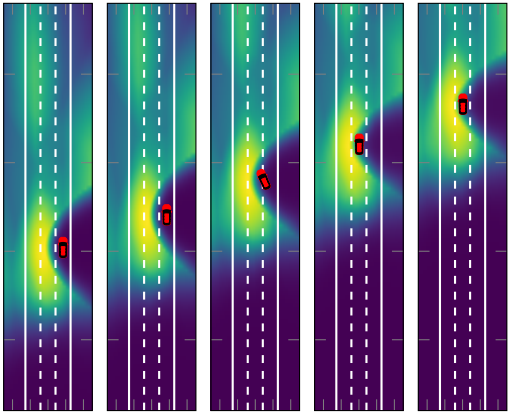

The heat map in \cref{fig:viz_2d} is symmetric across the front and back of the red vehicle. This symmetry is due to the fact that we are using the distance between the two cars as input to the neural network; the neural network does not have access to the $y-$position of the vehicle. The network therefore has no notion of whether or not the human-driven vehicle on the road is in front of or behind the robot-driven vehicle. To confirm this, we trained a new neural network with an additional input defined by $(y_r - y_h)/(v_r - v_h)$. Positive values mean the longitudinal distance between the two vehicles is increasing, while negative values indicate it is decreasing. \Cref{fig:viz_2d_time_gap} shows that when we add this fifth input to the network, we are able to break the symmetry.

While the user still prefers to be in the region of the road to the left of the other vehicle, it is now evident that the user would prefer being in front of the other vehicle to being behind it. The collision avoidance and staying on the road characteristics are preserved with the augmented input space. Adding the extra input did not have the same effect for all users, and future work should investigate the effect of changing the input space of the neural network.

\begin{figure}[htb]
    \centering
    \input{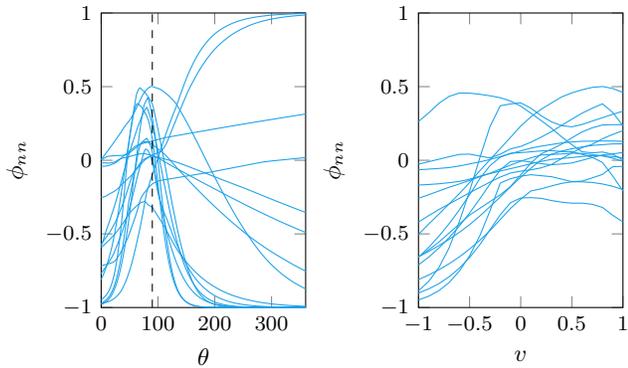}
    \caption{Variation in neural network feature value and hand-coded heading feature value with heading for all users. The vertical dashed line shows a heading of $90^\circ$, which corresponds to driving straight down the road. The remaining neural network inputs are held constant at $x_r=0.0$, $d=0.5$, $\theta_r = 90^\circ$ (right), and $v_r=0.8$ (left). \label{fig:all_users}}
\end{figure}

While similarities exist among the learned features, the features are unique to each particular user. \Cref{fig:all_users} shows the variation in the learned feature for each user with heading and speed.
The neural network feature favors driving straight down the road for most users, but it is less pronounced for some users and represents a trend that favors turning right, in which case the weight for the hand-coded heading feature is high enough to null out this trend. Similarly, most users have a negative feature value when driving backward, but the trend is more pronounced for some users.

\subsection{Customized Optimal Trajectories}
\begin{figure}[htb]
    \centering
    \input{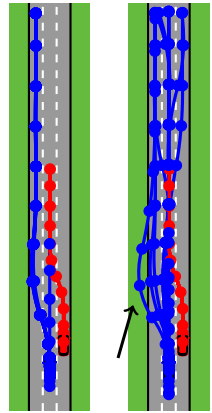}
    \caption{Optimized trajectories for all users when reward function is encoded with the hand-coded feature set (left) and with the mixed feature set (right). Marks indicate equal time intervals. The arrow points to user 3 for whom the training was unsuccessful.}
    \label{fig:opt_trajs} 
\end{figure}
Using the hand-coded driving features in previous work resulted in users converging to similar reward functions. While these reward functions resulted in safe driving, their optimal trajectories were almost indistinguishable from each other \cite{sadigh2017active}. \Cref{fig:opt_trajs} shows the optimal trajectories for the users in our study selected using only the hand-coded feature set and using the mixed feature set. Optimal trajectories were generated by selecting the trajectory that had the highest reward out of 10,000 trajectories. 

Consistent with the results of \citeauthor{sadigh2017active} \cite{sadigh2017active}, the optimal trajectories using the hand-coded feature set are almost identical among all users, except for user 13 (our further inspection shows that this user has chosen a large number of backward trajectories, hence their optimal trajectory obtained by the hand-coded features starts by moving backward).
In contrast, there is some variation among the optimal trajectories using the mixed feature set. Furthermore, with the exception of user 3 (in which the mixed feature set did not improve predictive accuracy over the hand-coded features), all users have safe trajectories that stay on the road and avoid collisions. These observations qualitatively support \textbf{H2}.

\section{Conclusion} \label{sec:disc}
In this work, we developed a method for learning extra features for a linear reward function based on user responses to preference queries. We applied our methods to a driving scenario used in previous work and found improvements in predictive power. Upon interpreting the neural network features of the users, we found that the neural network is able to represent a complex feature that is fine-tuned to user preferences. While different features were learned for different users, most users converged on features with some notion of speed, heading, and collision avoidance.

By providing a method for feature learning, we have taken a step towards overcoming the challenges of feature design. By learning an extra feature directly from users' preferences, we ensure that our features are representative enough to adequately model user reward. This benefit is reflected in the improvement in prediction accuracies over hand-coded only features on the test set. Moreover, the neural network feature is able to create a more expressive model that better distinguishes between the preferences of different users as evidenced by \cref{fig:opt_trajs}. Finally, our method relieves some of the burden of feature design by allowing the hand-coded features to be augmented with an additional learned feature.

This research provides multiple avenues for future work. In this work, queries shown to the users were based entirely on the hand-coded features, and the feature learning was performed offline after the data was collected. Future studies will explore an online approach, in which feature learning is interleaved with active querying of the user. In terms of interpretability, further analysis of the learned features and optimal trajectories could help to better understand and distinguish driving styles among users. While this study focused on learning features for individual users, future work will focus on collating the data to learn a universal feature that improves the prediction accuracy for all users. Finally, we note that the neural network learned a complex feature from the data that combined a number of aspects of driving such as maintaining heading and avoiding other vehicles. In this work, we used a mixed set of features to preserve some interpretability in the final reward function; however, we note based on the results presented here that a feature function of only neural network features may perform well and should be a subject of future work. Even though such an approach might be too data-hungry to avoid overfitting, it could be further extended beyond feature learning to the learning of nonlinear reward functions.

\section*{Acknowledgments}
The authors would like to acknowledge Dorsa Sadigh for her helpful input throughout the progression of this work. A.\,Maleki acknowledges the support of the Natural Sciences and Engineering Research Council of Canada (NSERC).

\renewcommand*{\bibfont}{\small}
\printbibliography

\end{document}

%% file: pastelcolormap.tex
\pgfplotsset{
   	colormap={pasteljet}{
		rgb=(0.99325,0.90616,0.14394)
		rgb=(0.98387,0.90487,0.13690)
		rgb=(0.97442,0.90359,0.13021)
		rgb=(0.96489,0.90232,0.12394)
		rgb=(0.95530,0.90107,0.11813)
		rgb=(0.94564,0.89982,0.11284)
		rgb=(0.93590,0.89857,0.10813)
		rgb=(0.92611,0.89733,0.10407)
		rgb=(0.91624,0.89609,0.10072)
		rgb=(0.90631,0.89485,0.09813)
		rgb=(0.89632,0.89362,0.09634)
		rgb=(0.88627,0.89237,0.09537)
		rgb=(0.87617,0.89112,0.09525)
		rgb=(0.86601,0.88987,0.09595)
		rgb=(0.85581,0.88860,0.09745)
		rgb=(0.84556,0.88732,0.09970)
		rgb=(0.83527,0.88603,0.10265)
		rgb=(0.82494,0.88472,0.10622)
		rgb=(0.81458,0.88339,0.11035)
		rgb=(0.80418,0.88205,0.11496)
		rgb=(0.79376,0.88068,0.12001)
		rgb=(0.78331,0.87928,0.12540)
		rgb=(0.77285,0.87787,0.13111)
		rgb=(0.76237,0.87642,0.13706)
		rgb=(0.75188,0.87495,0.14323)
		rgb=(0.74139,0.87345,0.14956)
		rgb=(0.73089,0.87192,0.15603)
		rgb=(0.72039,0.87035,0.16260)
		rgb=(0.70990,0.86875,0.16926)
		rgb=(0.69942,0.86712,0.17597)
		rgb=(0.68894,0.86545,0.18272)
		rgb=(0.67849,0.86374,0.18950)
		rgb=(0.66805,0.86200,0.19629)
		rgb=(0.65764,0.86022,0.20308)
		rgb=(0.64726,0.85840,0.20986)
		rgb=(0.63690,0.85654,0.21662)
		rgb=(0.62658,0.85464,0.22335)
		rgb=(0.61629,0.85271,0.23005)
		rgb=(0.60604,0.85073,0.23671)
		rgb=(0.59584,0.84872,0.24333)
		rgb=(0.58568,0.84666,0.24990)
		rgb=(0.57556,0.84457,0.25642)
		rgb=(0.56550,0.84243,0.26288)
		rgb=(0.55548,0.84025,0.26928)
		rgb=(0.54552,0.83804,0.27563)
		rgb=(0.53562,0.83579,0.28191)
		rgb=(0.52578,0.83349,0.28813)
		rgb=(0.51599,0.83116,0.29428)
		rgb=(0.50627,0.82879,0.30036)
		rgb=(0.49661,0.82638,0.30638)
		rgb=(0.48703,0.82393,0.31232)
		rgb=(0.47750,0.82144,0.31820)
		rgb=(0.46805,0.81892,0.32400)
		rgb=(0.45867,0.81636,0.32973)
		rgb=(0.44937,0.81377,0.33538)
		rgb=(0.44014,0.81114,0.34097)
		rgb=(0.43098,0.80847,0.34648)
		rgb=(0.42191,0.80577,0.35191)
		rgb=(0.41291,0.80304,0.35727)
		rgb=(0.40400,0.80027,0.36255)
		rgb=(0.39517,0.79748,0.36776)
		rgb=(0.38643,0.79464,0.37289)
		rgb=(0.37778,0.79178,0.37794)
		rgb=(0.36921,0.78889,0.38291)
		rgb=(0.36074,0.78596,0.38781)
		rgb=(0.35236,0.78301,0.39264)
		rgb=(0.34407,0.78003,0.39738)
		rgb=(0.33588,0.77702,0.40205)
		rgb=(0.32780,0.77398,0.40664)
		rgb=(0.31981,0.77091,0.41115)
		rgb=(0.31193,0.76782,0.41559)
		rgb=(0.30415,0.76470,0.41994)
		rgb=(0.29648,0.76156,0.42422)
		rgb=(0.28892,0.75839,0.42843)
		rgb=(0.28148,0.75520,0.43255)
		rgb=(0.27415,0.75199,0.43660)
		rgb=(0.26694,0.74875,0.44057)
		rgb=(0.25986,0.74549,0.44447)
		rgb=(0.25290,0.74221,0.44828)
		rgb=(0.24607,0.73891,0.45202)
		rgb=(0.23937,0.73559,0.45569)
		rgb=(0.23281,0.73225,0.45928)
		rgb=(0.22640,0.72889,0.46279)
		rgb=(0.22012,0.72551,0.46623)
		rgb=(0.21400,0.72211,0.46959)
		rgb=(0.20803,0.71870,0.47287)
		rgb=(0.20222,0.71527,0.47608)
		rgb=(0.19657,0.71183,0.47922)
		rgb=(0.19109,0.70837,0.48228)
		rgb=(0.18578,0.70489,0.48527)
		rgb=(0.18065,0.70140,0.48819)
		rgb=(0.17571,0.69790,0.49103)
		rgb=(0.17095,0.69438,0.49380)
		rgb=(0.16638,0.69086,0.49650)
		rgb=(0.16202,0.68732,0.49913)
		rgb=(0.15785,0.68376,0.50169)
		rgb=(0.15389,0.68020,0.50417)
		rgb=(0.15015,0.67663,0.50659)
		rgb=(0.14662,0.67305,0.50894)
		rgb=(0.14330,0.66946,0.51121)
		rgb=(0.14021,0.66586,0.51343)
		rgb=(0.13734,0.66225,0.51557)
		rgb=(0.13469,0.65864,0.51765)
		rgb=(0.13227,0.65501,0.51966)
		rgb=(0.13007,0.65138,0.52161)
		rgb=(0.12809,0.64775,0.52349)
		rgb=(0.12633,0.64411,0.52531)
		rgb=(0.12478,0.64046,0.52707)
		rgb=(0.12344,0.63681,0.52876)
		rgb=(0.12231,0.63315,0.53040)
		rgb=(0.12138,0.62949,0.53197)
		rgb=(0.12064,0.62583,0.53349)
		rgb=(0.12008,0.62216,0.53495)
		rgb=(0.11970,0.61849,0.53635)
		rgb=(0.11948,0.61482,0.53769)
		rgb=(0.11942,0.61114,0.53898)
		rgb=(0.11951,0.60746,0.54022)
		rgb=(0.11974,0.60379,0.54140)
		rgb=(0.12009,0.60010,0.54253)
		rgb=(0.12057,0.59642,0.54361)
		rgb=(0.12115,0.59274,0.54464)
		rgb=(0.12183,0.58905,0.54562)
		rgb=(0.12261,0.58537,0.54656)
		rgb=(0.12346,0.58169,0.54744)
		rgb=(0.12440,0.57800,0.54829)
		rgb=(0.12539,0.57432,0.54909)
		rgb=(0.12645,0.57063,0.54984)
		rgb=(0.12757,0.56695,0.55056)
		rgb=(0.12873,0.56327,0.55123)
		rgb=(0.12993,0.55958,0.55186)
		rgb=(0.13117,0.55590,0.55246)
		rgb=(0.13244,0.55222,0.55302)
		rgb=(0.13374,0.54853,0.55354)
		rgb=(0.13507,0.54485,0.55403)
		rgb=(0.13641,0.54117,0.55448)
		rgb=(0.13777,0.53749,0.55491)
		rgb=(0.13915,0.53381,0.55530)
		rgb=(0.14054,0.53013,0.55566)
		rgb=(0.14194,0.52645,0.55599)
		rgb=(0.14334,0.52277,0.55629)
		rgb=(0.14476,0.51909,0.55657)
		rgb=(0.14618,0.51541,0.55682)
		rgb=(0.14761,0.51173,0.55705)
		rgb=(0.14904,0.50805,0.55725)
		rgb=(0.15048,0.50437,0.55743)
		rgb=(0.15192,0.50069,0.55759)
		rgb=(0.15336,0.49700,0.55772)
		rgb=(0.15482,0.49331,0.55784)
		rgb=(0.15627,0.48962,0.55794)
		rgb=(0.15773,0.48593,0.55801)
		rgb=(0.15919,0.48224,0.55807)
		rgb=(0.16067,0.47854,0.55812)
		rgb=(0.16214,0.47484,0.55814)
		rgb=(0.16362,0.47113,0.55815)
		rgb=(0.16512,0.46742,0.55814)
		rgb=(0.16662,0.46371,0.55812)
		rgb=(0.16813,0.45999,0.55808)
		rgb=(0.16965,0.45626,0.55803)
		rgb=(0.17118,0.45253,0.55797)
		rgb=(0.17272,0.44879,0.55788)
		rgb=(0.17427,0.44504,0.55779)
		rgb=(0.17584,0.44129,0.55768)
		rgb=(0.17742,0.43753,0.55756)
		rgb=(0.17902,0.43376,0.55743)
		rgb=(0.18063,0.42997,0.55728)
		rgb=(0.18226,0.42618,0.55712)
		rgb=(0.18390,0.42238,0.55694)
		rgb=(0.18556,0.41857,0.55675)
		rgb=(0.18723,0.41475,0.55655)
		rgb=(0.18892,0.41091,0.55633)
		rgb=(0.19063,0.40706,0.55609)
		rgb=(0.19236,0.40320,0.55584)
		rgb=(0.19410,0.39932,0.55556)
		rgb=(0.19586,0.39543,0.55528)
		rgb=(0.19764,0.39153,0.55497)
		rgb=(0.19943,0.38761,0.55464)
		rgb=(0.20124,0.38367,0.55429)
		rgb=(0.20306,0.37972,0.55393)
		rgb=(0.20490,0.37575,0.55353)
		rgb=(0.20676,0.37176,0.55312)
		rgb=(0.20862,0.36775,0.55268)
		rgb=(0.21050,0.36373,0.55221)
		rgb=(0.21240,0.35968,0.55171)
		rgb=(0.21430,0.35562,0.55118)
		rgb=(0.21621,0.35153,0.55063)
		rgb=(0.21813,0.34743,0.55004)
		rgb=(0.22006,0.34331,0.54941)
		rgb=(0.22199,0.33916,0.54875)
		rgb=(0.22393,0.33499,0.54805)
		rgb=(0.22586,0.33081,0.54731)
		rgb=(0.22780,0.32659,0.54653)
		rgb=(0.22974,0.32236,0.54571)
		rgb=(0.23167,0.31811,0.54483)
		rgb=(0.23360,0.31383,0.54391)
		rgb=(0.23553,0.30953,0.54294)
		rgb=(0.23744,0.30520,0.54192)
		rgb=(0.23935,0.30085,0.54084)
		rgb=(0.24124,0.29648,0.53971)
		rgb=(0.24311,0.29209,0.53852)
		rgb=(0.24497,0.28768,0.53726)
		rgb=(0.24681,0.28324,0.53594)
		rgb=(0.24863,0.27877,0.53456)
		rgb=(0.25043,0.27429,0.53310)
		rgb=(0.25219,0.26978,0.53158)
		rgb=(0.25394,0.26525,0.52998)
		rgb=(0.25565,0.26070,0.52831)
		rgb=(0.25732,0.25613,0.52656)
		rgb=(0.25897,0.25154,0.52474)
		rgb=(0.26057,0.24692,0.52283)
		rgb=(0.26214,0.24229,0.52084)
		rgb=(0.26366,0.23763,0.51876)
		rgb=(0.26515,0.23296,0.51660)
		rgb=(0.26658,0.22826,0.51435)
		rgb=(0.26797,0.22355,0.51201)
		rgb=(0.26931,0.21882,0.50958)
		rgb=(0.27059,0.21407,0.50705)
		rgb=(0.27183,0.20930,0.50443)
		rgb=(0.27301,0.20452,0.50172)
		rgb=(0.27413,0.19972,0.49891)
		rgb=(0.27519,0.19490,0.49600)
		rgb=(0.27619,0.19007,0.49300)
		rgb=(0.27713,0.18523,0.48990)
		rgb=(0.27801,0.18037,0.48670)
		rgb=(0.27883,0.17549,0.48340)
		rgb=(0.27957,0.17060,0.48000)
		rgb=(0.28025,0.16569,0.47650)
		rgb=(0.28087,0.16077,0.47290)
		rgb=(0.28141,0.15583,0.46920)
		rgb=(0.28189,0.15088,0.46541)
		rgb=(0.28229,0.14591,0.46151)
		rgb=(0.28262,0.14093,0.45752)
		rgb=(0.28288,0.13592,0.45343)
		rgb=(0.28307,0.13090,0.44924)
		rgb=(0.28319,0.12585,0.44496)
		rgb=(0.28323,0.12078,0.44058)
		rgb=(0.28320,0.11568,0.43611)
		rgb=(0.28309,0.11055,0.43155)
		rgb=(0.28291,0.10539,0.42690)
		rgb=(0.28266,0.10020,0.42216)
		rgb=(0.28233,0.09495,0.41733)
		rgb=(0.28192,0.08967,0.41241)
		rgb=(0.28145,0.08432,0.40741)
		rgb=(0.28089,0.07891,0.40233)
		rgb=(0.28027,0.07342,0.39716)
		rgb=(0.27957,0.06784,0.39192)
		rgb=(0.27879,0.06214,0.38659)
		rgb=(0.27794,0.05632,0.38119)
		rgb=(0.27702,0.05034,0.37572)
		rgb=(0.27602,0.04417,0.37016)
		rgb=(0.27495,0.03775,0.36454)
		rgb=(0.27381,0.03150,0.35885)
		rgb=(0.27259,0.02556,0.35309)
		rgb=(0.27131,0.01994,0.34727)
		rgb=(0.26994,0.01463,0.34138)
		rgb=(0.26851,0.00961,0.33543)
		rgb=(0.26700,0.00487,0.32942)
	  }
}

%% file: pastelcolorlist.tex
\pgfplotscreateplotcyclelist{pastelcolors}{%
  solid, pastelPurple, mark=none\\%
  solid, pastelBlue, mark=none\\%
  solid, pastelGreen, mark=none\\%
  solid, pastelRed, mark=none\\%
  solid, pastelMagenta, mark=none\\%
  solid, pastelOrange, mark=none\\%
  solid, pastelSeaGreen, mark=none\\%
}

%% file: ig_hs_traj.tex

\hspace{-5cm}
\begin{tikzpicture}   
\node at (1.2, 6.0) {Information Gain};
\begin{groupplot}[group style={horizontal sep = 0.2cm, group size=2 by 1}]
\nextgroupplot [height = {7cm}, xmin = {-0.5}, xmax = {0.5}, axis equal image = {true}, ymax = {4}, hide axis = {true}, ymin = {-0.5}, width = {3cm}]\addplot+ [mark = {none}, black,solid,thick, name path=A]coordinates {
(-0.255, -1.0)
(-0.255, 5.0)
};
\addplot+ [mark = {none}, black,solid,thick, name path=B]coordinates {
(0.255, -1.0)
(0.255, 5.0)
};
\addplot+ [mark = {none}, white,dashed,thick]coordinates {
(-0.085, -1.0)
(-0.085, 5.0)
};
\addplot+ [mark = {none}, white,dashed,thick]coordinates {
(0.085, -1.0)
(0.085, 5.0)
};
\path[name path=axis] (axis cs:-0.5,-1) -- (axis cs:-0.5,5);;
\path[name path=axisright] (axis cs:0.5,-1) -- (axis cs:0.5,5);;
\addplot[pastelGreen!80] fill between[of=A and axis];;
\addplot[pastelGreen!80] fill between[of=B and axisright];;
\addplot[black!40] fill between[of=A and B];;
\addplot+ [mark = {none}, blue,solid, very thick]coordinates {
(2.4492935982947068e-18, -0.26)
(0.0010487091343298147, -0.2140119558020612)
(0.0035673281611791107, -0.16267369951888)
(0.007968995860849256, -0.10658615258142087)
(0.014648568178484686, -0.04632119369440695)
(0.023975181776995844, 0.017572285340186768)
(0.036286962989891855, 0.0845638846247044)
(0.051887396879017035, 0.15413850780621816)
(0.07104298606975151, 0.22579424135729192)
(0.09398191490586015, 0.29904108677005453)
(0.12089349871316354, 0.37340035876964134)
(0.13993697153309106, 0.4427028777707514)
(0.15268945588349825, 0.5068314313904275)
(0.16046253693679405, 0.5658677092847254)
(0.16433386873032788, 0.620020731829991)
(0.16518015557243998, 0.6695756080845664)
(0.16370860765035497, 0.7148575955030448)
(0.16048541896487822, 0.7562074659636887)
(0.15596069851751743, 0.7939650927363793)
(0.15048979806872229, 0.8284589260399002)
(0.14435125651628442, 0.85999962331429)
(0.13737006959578069, 0.8939076640851112)
(0.129516907214241, 0.9299186344589023)
(0.12076348345006474, 0.9677918179740839)
(0.11108336408820993, 1.0073078674475668)
(0.1004524912904062, 1.0482667166708355)
(0.08884949526220497, 1.090485707305865)
(0.07625584758227884, 1.133797908149407)
(0.06265589878024865, 1.1780506060016382)
(0.048036833166622064, 1.2231039494859366)
(0.03238856633234718, 1.2688297291990487)
(0.01902252658561078, 1.3146774605907672)
(0.007877861979743668, 1.3605847422848896)
(-0.0011003730406338192, 1.4064916362942903) 
(-0.007962104376642881, 1.4523399637789023)
(-0.012753051883033991, 1.498072757131174)
(-0.01551547129857709, 1.5436338413067314)
(-0.016288827189116978, 1.5889675218498855)
(-0.015110400360418916, 1.6340183608395704)
(-0.012015833628624031, 1.6787310251234793)
(-0.007039620030108906, 1.723050193821023)
(-0.0022325035569350496, 1.7729571090326988)
(0.002078018798493228, 1.82791242218332)
(0.005578945174934086, 1.8874212259652212)
(0.00797546525414495, 1.9510266197534036)
(0.00899331013137485, 2.0183043938799115)
(0.008380092082775702, 2.0888586549089725)
(0.005905892736829173, 2.162318233109378)
(0.0013632982675150732, 2.238333734032815)
(-0.005432965848518026, 2.3165751164628734)
};
\addplot+ [mark = {none}, red,solid, very thick]coordinates {
(0.17, 0.041)
(0.17, 0.082)
(0.17, 0.123)
(0.17, 0.164)
(0.17, 0.20500000000000002)
(0.17, 0.24600000000000002)
(0.17, 0.28700000000000003)
(0.17, 0.328)
(0.17, 0.369)
(0.17, 0.41)
(0.17, 0.45099999999999996)
(0.16831947092058408, 0.4919655443270712)
(0.16496123733542573, 0.5328277795472292)
(0.1599309436444288, 0.5735180258647721)
(0.15323704458682028, 0.6139678925568459)
(0.1448907910307244, 0.6541093929221456)
(0.13490621106310258, 0.6938750585505513)
(0.12330008641184245, 0.7331980527216392)
(0.1100919242396247, 0.7720122827414708)
(0.09530392435697597, 0.8102525110288488)
(0.07896094190961467, 0.8478544647643296)
(0.06417294202696594, 0.8860946930517075)
(0.05096477985474819, 0.9249089230715392)
(0.039358655203488066, 0.964231917242627)
(0.02937407523586627, 1.0039975828710328)
(0.021027821679770388, 1.0441390832363324)
(0.014333922622161863, 1.0845889499284063)
(0.009303628931164945, 1.1252791962459492)
(0.005945395346006596, 1.1661414314661072)
(0.0042648662665906506, 1.2071069757931783)
(0.004264866266590653, 1.2481069757931782)
(0.004264866266590656, 1.2903369757931782)
(0.004264866266590658, 1.3336739757931781)
(0.004264866266590661, 1.378007275793178)
(0.004264866266590664, 1.4232372457931781)
(0.004264866266590666, 1.4692742187931782)
(0.004264866266590669, 1.5160374944931783)
(0.004264866266590671, 1.5634544426231782)
(0.004264866266590674, 1.6114596959401783)
(0.004264866266590677, 1.6599944239254785)
(0.004264866266590679, 1.7090056791122485)
(0.004264866266590682, 1.7584458087803416)
(0.004264866266590685, 1.8082719254816253)
(0.004264866266590689, 1.8584454305127807)
(0.004264866266590692, 1.9089315850408204)
(0.004264866266590696, 1.9596991241160562)
(0.004264866266590699, 2.010719909283768)
(0.004264866266590703, 2.0619686159347093)
(0.004264866266590706, 2.113422451920556)
(0.0042648662665907095, 2.1650609043078184)
};
\node[sedan top,body color=red,window color=black, minimum width=1.5cm,rotate=90.0,scale = 0.27] at (axis cs:0.17, 0.041) {};;
\node[sedan top,body color=blue,window color=black, minimum width=1.5cm,rotate=88.69365622710173,scale = 0.27] at (axis cs:2.4492935982947068e-18, -0.26) {};;
\addplot+[scatter, scatter src=explicit symbolic, only marks = {true}, scatter/classes = {{a={mark=*,red},b={mark=*,blue}}}] coordinates {
(0.17, 0.041) [a]
(0.17, 0.24600000000000002) [a]
(0.17, 0.45099999999999996) [a]
(0.1448907910307244, 0.6541093929221456) [a]
(0.07896094190961467, 0.8478544647643296) [a]
(0.021027821679770388, 1.0441390832363324) [a]
(0.004264866266590653, 1.2481069757931782) [a]
(0.004264866266590666, 1.4692742187931782) [a]
(0.004264866266590679, 1.7090056791122485) [a]
(0.004264866266590696, 1.9596991241160562) [a]
(0.0042648662665907095, 2.1650609043078184) [a]
};
\addplot+[scatter, scatter src=explicit symbolic, only marks = {true}, scatter/classes = {{a={mark=*,red},b={mark=*,blue}}}] coordinates {
(2.4492935982947068e-18, -0.26) [b]
(0.023975181776995844, 0.017572285340186768) [b]
(0.12089349871316354, 0.37340035876964134) [b]
(0.16518015557243998, 0.6695756080845664) [b]
(0.14435125651628442, 0.85999962331429) [b]
(0.1004524912904062, 1.0482667166708355) [b]
(0.03238856633234718, 1.2688297291990487) [b]
(-0.012753051883033991, 1.498072757131174) [b]
(-0.007039620030108906, 1.723050193821023) [b]
(0.00899331013137485, 2.0183043938799115) [b]
(-0.005432965848518026, 2.3165751164628734) [b]
};
\nextgroupplot [height = {7cm}, xmin = {-0.5}, xmax = {0.5}, axis equal image = {true}, ymax = {4}, hide axis = {true}, ymin = {-0.5}, width = {3cm}]\addplot+ [mark = {none}, black,solid,thick, name path=A]coordinates {
(-0.255, -1.0)
(-0.255, 5.0)
};
\addplot+ [mark = {none}, black,solid,thick, name path=B]coordinates {
(0.255, -1.0)
(0.255, 5.0)
};
\addplot+ [mark = {none}, white,dashed,thick]coordinates {
(-0.085, -1.0)
(-0.085, 5.0)
};
\addplot+ [mark = {none}, white,dashed,thick]coordinates {
(0.085, -1.0)
(0.085, 5.0)
};
\path[name path=axis] (axis cs:-0.5,-1) -- (axis cs:-0.5,5);;
\path[name path=axisright] (axis cs:0.5,-1) -- (axis cs:0.5,5);;
\addplot[pastelGreen!80] fill between[of=A and axis];;
\addplot[pastelGreen!80] fill between[of=B and axisright];;
\addplot[black!40] fill between[of=A and B];;
\addplot+ [mark = {none}, blue,solid, very thick]coordinates {
(2.4492935982947068e-18, -0.26)
(0.0018395093725851782, -0.21403679509359502)
(0.006254462507492783, -0.16282675517155853)
(0.01396028693017104, -0.10709697973717466)
(0.02562940679534001, -0.047596445306193974)
(0.04187292740728129, 0.014897637793643878)
(0.0632257454437478, 0.07957770808761581)
(0.09013460132830586, 0.14560736205734887)
(0.12294876427919874, 0.21212586668439917)
(0.16191314515724586, 0.2782550723492162)
(0.20716368911211958, 0.34310815976310954)
(0.25872492545936177, 0.4057997004847247)
(0.3165095580439038, 0.46545656089804366)
(0.3803199739359552, 0.5212292314249316)
(0.4498515374138974, 0.5723032143673987)
(0.5246975236187195, 0.6179101539016689)
(0.6043555343993321, 0.6573384395527612)
(0.6882352290821923, 0.6899430594117613)
(0.7756671958676857, 0.7151545210674874)
(0.8659127854886401, 0.7324866965175493)
(0.9581747275761783, 0.7415434821359729)
(1.030280353217297, 0.7554523825201707)
(1.0844285196267347, 0.7700885834460799)
(1.1228546489475597, 0.7828260913783143)
(1.1475889002935789, 0.7921520699112533)
(1.1603596432747436, 0.7973567894119008)
(1.1625800872274967, 0.7982976434038813)
(1.1553783390391081, 0.7952255878582021)
(1.1396467786573186, 0.7886600663453593)
(1.1160968587413924, 0.7792996607562076)
(1.0853118094525445, 0.7679580883037725)
(1.0486899825547078, 0.7530844274623815)
(1.0071759850925068, 0.7342800367975033)
(0.9617115693438943, 0.7111346657623319)
(0.9132404007782019, 0.6832589813907695)
(0.8627100670632387, 0.6503100791177917)
(0.8110712303317907, 0.6120112302855975)
(0.7592741301147183, 0.5681667944913984)
(0.7082628125673535, 0.5186729908441846)
(0.6589675418930753, 0.4635250602294655)
(0.6122958742765557, 0.40282124295057514)
(0.569122864142857, 0.3367639271920747)
(0.5302808409058373, 0.26565828127509544)
(0.496549150630228, 0.18990865725227674)
(0.46864420722585215, 0.11001303818907998)
(0.44721014592563746, 0.02655579173907527)
(0.43281032052372237, -0.05980101481511453)
(0.4259198368893937, -0.1483274899840391)
(0.4269192696992258, -0.238236743765404)
(0.43608966775291913, -0.3286964326047352)
};
\addplot+ [mark = {none}, red,solid, very thick]coordinates {
(0.17, 0.041)
(0.17, 0.082)
(0.17, 0.123)
(0.17, 0.164)
(0.17, 0.20500000000000002)
(0.17, 0.24600000000000002)
(0.17, 0.28700000000000003)
(0.17, 0.328)
(0.17, 0.369)
(0.17, 0.41)
(0.17, 0.45099999999999996)
(0.16831947092058408, 0.4919655443270712)
(0.16496123733542573, 0.5328277795472292)
(0.1599309436444288, 0.5735180258647721)
(0.15323704458682028, 0.6139678925568459)
(0.1448907910307244, 0.6541093929221456)
(0.13490621106310258, 0.6938750585505513)
(0.12330008641184245, 0.7331980527216392)
(0.1100919242396247, 0.7720122827414708)
(0.09530392435697597, 0.8102525110288488)
(0.07896094190961467, 0.8478544647643296)
(0.06417294202696594, 0.8860946930517075)
(0.05096477985474819, 0.9249089230715392)
(0.039358655203488066, 0.964231917242627)
(0.02937407523586627, 1.0039975828710328)
(0.021027821679770388, 1.0441390832363324)
(0.014333922622161863, 1.0845889499284063)
(0.009303628931164945, 1.1252791962459492)
(0.005945395346006596, 1.1661414314661072)
(0.0042648662665906506, 1.2071069757931783)
(0.004264866266590653, 1.2481069757931782)
(0.004264866266590656, 1.2903369757931782)
(0.004264866266590658, 1.3336739757931781)
(0.004264866266590661, 1.378007275793178)
(0.004264866266590664, 1.4232372457931781)
(0.004264866266590666, 1.4692742187931782)
(0.004264866266590669, 1.5160374944931783)
(0.004264866266590671, 1.5634544426231782)
(0.004264866266590674, 1.6114596959401783)
(0.004264866266590677, 1.6599944239254785)
(0.004264866266590679, 1.7090056791122485)
(0.004264866266590682, 1.7584458087803416)
(0.004264866266590685, 1.8082719254816253)
(0.004264866266590689, 1.8584454305127807)
(0.004264866266590692, 1.9089315850408204)
(0.004264866266590696, 1.9596991241160562)
(0.004264866266590699, 2.010719909283768)
(0.004264866266590703, 2.0619686159347093)
(0.004264866266590706, 2.113422451920556)
(0.0042648662665907095, 2.1650609043078184)
};
\node[sedan top,body color=red,window color=black, minimum width=1.5cm,rotate=90.0,scale = 0.27] at (axis cs:0.17, 0.041) {};;
\node[sedan top,body color=blue,window color=black, minimum width=1.5cm,rotate=87.70816881947671,scale = 0.27] at (axis cs:2.4492935982947068e-18, -0.26) {};;
\addplot+[scatter, scatter src=explicit symbolic, only marks = {true}, scatter/classes = {{a={mark=*,red},b={mark=*,blue}}}] coordinates {
(0.17, 0.041) [a]
(0.17, 0.24600000000000002) [a]
(0.17, 0.45099999999999996) [a]
(0.1448907910307244, 0.6541093929221456) [a]
(0.07896094190961467, 0.8478544647643296) [a]
(0.021027821679770388, 1.0441390832363324) [a]
(0.004264866266590653, 1.2481069757931782) [a]
(0.004264866266590666, 1.4692742187931782) [a]
(0.004264866266590679, 1.7090056791122485) [a]
(0.004264866266590696, 1.9596991241160562) [a]
(0.0042648662665907095, 2.1650609043078184) [a]
};
\addplot+[scatter, scatter src=explicit symbolic, only marks = {true}, scatter/classes = {{a={mark=*,red},b={mark=*,blue}}}] coordinates {
(2.4492935982947068e-18, -0.26) [b]
(0.04187292740728129, 0.014897637793643878) [b]
(0.20716368911211958, 0.34310815976310954) [b]
(0.5246975236187195, 0.6179101539016689) [b]
(0.9581747275761783, 0.7415434821359729) [b]
(1.1603596432747436, 0.7973567894119008) [b]
(1.0853118094525445, 0.7679580883037725) [b]
(0.8627100670632387, 0.6503100791177917) [b]
(0.6122958742765557, 0.40282124295057514) [b]
(0.44721014592563746, 0.02655579173907527) [b]
(0.43608966775291913, -0.3286964326047352) [b]
};
\end{groupplot}  
\end{tikzpicture}%

\vspace{-6.21cm}

\hspace{5cm}
\begin{tikzpicture}

\node at (1.2, 6.0) {Heuristic};
\begin{groupplot}[group style={horizontal sep = 0.2cm, group size=2 by 1}]
\nextgroupplot [height = {7cm}, xmin = {-0.5}, xmax = {0.5}, axis equal image = {true}, ymax = {4}, hide axis = {true}, ymin = {-0.5}, width = {3cm}]\addplot+ [mark = {none}, black,solid,thick, name path=A]coordinates {
(-0.255, -1.0)
(-0.255, 5.0)
};
\addplot+ [mark = {none}, black,solid,thick, name path=B]coordinates {
(0.255, -1.0)
(0.255, 5.0)
};
\addplot+ [mark = {none}, white,dashed,thick]coordinates {
(-0.085, -1.0)
(-0.085, 5.0)
};
\addplot+ [mark = {none}, white,dashed,thick]coordinates {
(0.085, -1.0)
(0.085, 5.0)
};
\path[name path=axis] (axis cs:-0.5,-1) -- (axis cs:-0.5,5);;
\path[name path=axisright] (axis cs:0.5,-1) -- (axis cs:0.5,5);;
\addplot[pastelGreen!80] fill between[of=A and axis];;
\addplot[pastelGreen!80] fill between[of=B and axisright];;
\addplot[black!40] fill between[of=A and B];;
\addplot+ [mark = {none}, blue,solid, very thick]coordinates {
(2.4492935982947068e-18, -0.26)
(-8.839982968276287e-5, -0.23400015027985524)
(-0.00016357343536944322, -0.22060036114237222)
(-0.0001774762698255124, -0.21854040805766517)
(-0.00012107300191985598, -0.22668621278595136)
(-1.3069788889102365e-5, -0.24401727626359102)
(0.00010873960918417712, -0.2696152464467548)
(0.0001940668503226182, -0.3026535702606875)
(0.00018510280662582427, -0.3423881598495524)
(2.0224058763838578e-5, -0.38814899435701833)
(-0.00036328809933606235, -0.43933257595893194)
(-0.0005445789081618016, -0.4785186731461014)
(-0.0006136160954024411, -0.5069064540919713)
(-0.0006293393403848675, -0.5255755258742437)
(-0.0006273225559524424, -0.5354976962323321)
(-0.0006257668423666076, -0.5375476491487642)
(-0.0006301659546277236, -0.5325126092265893)
(-0.0006369185834488679, -0.5211010735649464)
(-0.0006361072143025123, -0.5039506896905591)
(-0.0006136193617423039, -0.48163535551718667)
(-0.0005527515778027941, -0.4546716132645778)
(-0.0004919454228608498, -0.4260242479388694)
(-0.00043224343810181114, -0.39586162015138726)
(-0.0003745967744866344, -0.3643352546701294)
(-0.00031986890111381714, -0.3315815240254313)
(-0.00026884027320815576, -0.29772316374844515)
(-0.0002222135811210953, -0.2628706360805832)
(-0.0001806192962795086, -0.227123357308015)
(-0.00014462130404674188, -0.19057080235968268)
(-0.00011472247114735806, -0.15329349894326486)
(-9.137004001378227e-5, -0.11536392226591882)
(-6.934183220564497e-5, -0.07455730273201819)
(-4.945753499654827e-5, -0.031161344355996927)
(-3.2474111551743714e-5, 0.014565019128477998)
(-1.9085308097924795e-5, 0.062388747228879056)
(-9.922806641760219e-6, 0.11210010336160847)
(-5.558524304522548e-6, 0.16351032445577007)
(-6.5076768739549465e-6, 0.21644952359872716)
(-1.3232316082350961e-5, 0.270764802418766)
(-2.6145122360236412e-5, 0.32631855223073947)
(-4.56132913731978e-5, 0.3829869250680564)
(-6.311324111918971e-5, 0.4439884611211649)
(-7.777288356278241e-5, 0.5088898441724858)
(-8.878533373359047e-5, 0.5773010895223795)
(-9.54100803779907e-5, 0.6488712108284049)
(-9.697222728112283e-5, 0.7232843202633737)
(-9.286037762072956e-5, 0.8002561186597751)
(-8.252360087721659e-5, 0.8795307366414649)
(-6.546781698548237e-5, 0.9608778916434962)
(-4.125184994383774e-5, 1.0440903292309471)
};
\addplot+ [mark = {none}, red,solid, very thick]coordinates {
(0.17, 0.041)
(0.17, 0.082)
(0.17, 0.123)
(0.17, 0.164)
(0.17, 0.20500000000000002)
(0.17, 0.24600000000000002)
(0.17, 0.28700000000000003)
(0.17, 0.328)
(0.17, 0.369)
(0.17, 0.41)
(0.17, 0.45099999999999996)
(0.16831947092058408, 0.4919655443270712)
(0.16496123733542573, 0.5328277795472292)
(0.1599309436444288, 0.5735180258647721)
(0.15323704458682028, 0.6139678925568459)
(0.1448907910307244, 0.6541093929221456)
(0.13490621106310258, 0.6938750585505513)
(0.12330008641184245, 0.7331980527216392)
(0.1100919242396247, 0.7720122827414708)
(0.09530392435697597, 0.8102525110288488)
(0.07896094190961467, 0.8478544647643296)
(0.06417294202696594, 0.8860946930517075)
(0.05096477985474819, 0.9249089230715392)
(0.039358655203488066, 0.964231917242627)
(0.02937407523586627, 1.0039975828710328)
(0.021027821679770388, 1.0441390832363324)
(0.014333922622161863, 1.0845889499284063)
(0.009303628931164945, 1.1252791962459492)
(0.005945395346006596, 1.1661414314661072)
(0.0042648662665906506, 1.2071069757931783)
(0.004264866266590653, 1.2481069757931782)
(0.004264866266590656, 1.2903369757931782)
(0.004264866266590658, 1.3336739757931781)
(0.004264866266590661, 1.378007275793178)
(0.004264866266590664, 1.4232372457931781)
(0.004264866266590666, 1.4692742187931782)
(0.004264866266590669, 1.5160374944931783)
(0.004264866266590671, 1.5634544426231782)
(0.004264866266590674, 1.6114596959401783)
(0.004264866266590677, 1.6599944239254785)
(0.004264866266590679, 1.7090056791122485)
(0.004264866266590682, 1.7584458087803416)
(0.004264866266590685, 1.8082719254816253)
(0.004264866266590689, 1.8584454305127807)
(0.004264866266590692, 1.9089315850408204)
(0.004264866266590696, 1.9596991241160562)
(0.004264866266590699, 2.010719909283768)
(0.004264866266590703, 2.0619686159347093)
(0.004264866266590706, 2.113422451920556)
(0.0042648662665907095, 2.1650609043078184)
};
\node[sedan top,body color=red,window color=black, minimum width=1.5cm,rotate=90.0,scale = 0.27] at (axis cs:0.17, 0.041) {};;
\node[sedan top,body color=blue,window color=black, minimum width=1.5cm,rotate=90.19480565034448,scale = 0.27] at (axis cs:2.4492935982947068e-18, -0.26) {};;
\addplot+[scatter, scatter src=explicit symbolic, only marks = {true}, scatter/classes = {{a={mark=*,red},b={mark=*,blue}}}] coordinates {
(0.17, 0.041) [a]
(0.17, 0.24600000000000002) [a]
(0.17, 0.45099999999999996) [a]
(0.1448907910307244, 0.6541093929221456) [a]
(0.07896094190961467, 0.8478544647643296) [a]
(0.021027821679770388, 1.0441390832363324) [a]
(0.004264866266590653, 1.2481069757931782) [a]
(0.004264866266590666, 1.4692742187931782) [a]
(0.004264866266590679, 1.7090056791122485) [a]
(0.004264866266590696, 1.9596991241160562) [a]
(0.0042648662665907095, 2.1650609043078184) [a]
};
\addplot+[scatter, scatter src=explicit symbolic, only marks = {true}, scatter/classes = {{a={mark=*,red},b={mark=*,blue}}}] coordinates {
(2.4492935982947068e-18, -0.26) [b]
(-1.3069788889102365e-5, -0.24401727626359102) [b]
(-0.00036328809933606235, -0.43933257595893194) [b]
(-0.0006257668423666076, -0.5375476491487642) [b]
(-0.0005527515778027941, -0.4546716132645778) [b]
(-0.00026884027320815576, -0.29772316374844515) [b]
(-9.137004001378227e-5, -0.11536392226591882) [b]
(-9.922806641760219e-6, 0.11210010336160847) [b]
(-4.56132913731978e-5, 0.3829869250680564) [b]
(-9.697222728112283e-5, 0.7232843202633737) [b]
(-4.125184994383774e-5, 1.0440903292309471) [b]
};
\nextgroupplot [height = {7cm}, xmin = {-0.5}, xmax = {0.5}, axis equal image = {true}, ymax = {4}, hide axis = {true}, ymin = {-0.5}, width = {3cm}]\addplot+ [mark = {none}, black,solid,thick, name path=A]coordinates {
(-0.255, -1.0)
(-0.255, 5.0)
};
\addplot+ [mark = {none}, black,solid,thick, name path=B]coordinates {
(0.255, -1.0)
(0.255, 5.0)
};
\addplot+ [mark = {none}, white,dashed,thick]coordinates {
(-0.085, -1.0)
(-0.085, 5.0)
};
\addplot+ [mark = {none}, white,dashed,thick]coordinates {
(0.085, -1.0)
(0.085, 5.0)
};
\path[name path=axis] (axis cs:-0.5,-1) -- (axis cs:-0.5,5);;
\path[name path=axisright] (axis cs:0.5,-1) -- (axis cs:0.5,5);;
\addplot[pastelGreen!80] fill between[of=A and axis];;
\addplot[pastelGreen!80] fill between[of=B and axisright];;
\addplot[black!40] fill between[of=A and B];;
\addplot+ [mark = {none}, blue,solid, very thick]coordinates {
(2.4492935982947068e-18, -0.26)
(-0.0010376719736254318, -0.21401170543632703)
(-0.0035298001261094393, -0.16267215638034488)
(-0.007885227761557478, -0.10658099976798417)
(-0.01449477011410678, -0.04630832017026048)
(-0.02372388652432624, 0.017599314699499942)
(-0.03590747778603699, 0.08461434611260282)
(-0.05134632661915576, 0.15422500388198973)
(-0.07030481316531073, 0.225933138259241)
(-0.09300962263501689, 0.2992528930734351)
(-0.11964922610498104, 0.3737100364667263)
(-0.1432335097124717, 0.45137966977619637)
(-0.16335612563228505, 0.5319593513705563)
(-0.1796506857486075, 0.6151269367907962)
(-0.1917907290359947, 0.7005424471052285)
(-0.19948953853459253, 0.787850195426665)
(-0.2024998128600484, 0.8766810832945335)
(-0.20061319878776324, 0.9666549967560586)
(-0.19365969280299866, 1.0573832468731947)
(-0.18150692049397862, 1.1484710115678591)
(-0.16405930325910698, 1.239519745641034)
(-0.14505587217450652, 1.3310016716816018)
(-0.12448382342583637, 1.4228165810478608)
(-0.10233343312562847, 1.514871381233135)
(-0.07859774268894941, 1.6070794359255518)
(-0.053272275009925624, 1.6993599685007101)
(-0.02635477935668433, 1.7916375227029921)
(0.002154997164971771, 1.8838414749476624)
(0.03225551373014334, 1.9759055932658043)
(0.06394361760035872, 2.0677676384307415)
(0.0972147056410751, 2.159369003258805)
(0.12469181695927439, 2.253136929151775)
(0.14621243604508913, 2.3486830717331277)
(0.16164341085526301, 2.4456082242047077)
(0.1708812902730812, 2.5435045587171192)
(0.1738525267036508, 2.641957840922275)
(0.17051354726580212, 2.7405496130201747)
(0.16085069700613433, 2.838859340736785)
(0.14488005753874386, 2.9364665197458995)
(0.12264714451455612, 3.0329527370881704)
(0.0942264873522882, 3.1279036831570197)
(0.0693695448882963, 3.223940842357081)
(0.04812299381970843, 3.320922438103544)
(0.03052709732948923, 3.4187053570896286)
(0.01661577028939119, 3.5171453931440726)
(0.006416642694785807, 3.616097484945837)
(-4.88779532305892e-5, 3.7154159483318643)
(-0.002765544643083746, 3.8149547038196814)
(-0.0017242085910582812, 3.9145674998626134)
(0.003078243869734422, 4.014108132260943)
};
\addplot+ [mark = {none}, red,solid, very thick]coordinates {
(0.17, 0.041)
(0.17, 0.082)
(0.17, 0.123)
(0.17, 0.164)
(0.17, 0.20500000000000002)
(0.17, 0.24600000000000002)
(0.17, 0.28700000000000003)
(0.17, 0.328)
(0.17, 0.369)
(0.17, 0.41)
(0.17, 0.45099999999999996)
(0.16831947092058408, 0.4919655443270712)
(0.16496123733542573, 0.5328277795472292)
(0.1599309436444288, 0.5735180258647721)
(0.15323704458682028, 0.6139678925568459)
(0.1448907910307244, 0.6541093929221456)
(0.13490621106310258, 0.6938750585505513)
(0.12330008641184245, 0.7331980527216392)
(0.1100919242396247, 0.7720122827414708)
(0.09530392435697597, 0.8102525110288488)
(0.07896094190961467, 0.8478544647643296)
(0.06417294202696594, 0.8860946930517075)
(0.05096477985474819, 0.9249089230715392)
(0.039358655203488066, 0.964231917242627)
(0.02937407523586627, 1.0039975828710328)
(0.021027821679770388, 1.0441390832363324)
(0.014333922622161863, 1.0845889499284063)
(0.009303628931164945, 1.1252791962459492)
(0.005945395346006596, 1.1661414314661072)
(0.0042648662665906506, 1.2071069757931783)
(0.004264866266590653, 1.2481069757931782)
(0.004264866266590656, 1.2903369757931782)
(0.004264866266590658, 1.3336739757931781)
(0.004264866266590661, 1.378007275793178)
(0.004264866266590664, 1.4232372457931781)
(0.004264866266590666, 1.4692742187931782)
(0.004264866266590669, 1.5160374944931783)
(0.004264866266590671, 1.5634544426231782)
(0.004264866266590674, 1.6114596959401783)
(0.004264866266590677, 1.6599944239254785)
(0.004264866266590679, 1.7090056791122485)
(0.004264866266590682, 1.7584458087803416)
(0.004264866266590685, 1.8082719254816253)
(0.004264866266590689, 1.8584454305127807)
(0.004264866266590692, 1.9089315850408204)
(0.004264866266590696, 1.9596991241160562)
(0.004264866266590699, 2.010719909283768)
(0.004264866266590703, 2.0619686159347093)
(0.004264866266590706, 2.113422451920556)
(0.0042648662665907095, 2.1650609043078184)
};
\node[sedan top,body color=red,window color=black, minimum width=1.5cm,rotate=90.0,scale = 0.27] at (axis cs:0.17, 0.041) {};;
\node[sedan top,body color=blue,window color=black, minimum width=1.5cm,rotate=91.29259278581513,scale = 0.27] at (axis cs:2.4492935982947068e-18, -0.26) {};;
\addplot+[scatter, scatter src=explicit symbolic, only marks = {true}, scatter/classes = {{a={mark=*,red},b={mark=*,blue}}}] coordinates {
(0.17, 0.041) [a]
(0.17, 0.24600000000000002) [a]
(0.17, 0.45099999999999996) [a]
(0.1448907910307244, 0.6541093929221456) [a]
(0.07896094190961467, 0.8478544647643296) [a]
(0.021027821679770388, 1.0441390832363324) [a]
(0.004264866266590653, 1.2481069757931782) [a]
(0.004264866266590666, 1.4692742187931782) [a]
(0.004264866266590679, 1.7090056791122485) [a]
(0.004264866266590696, 1.9596991241160562) [a]
(0.0042648662665907095, 2.1650609043078184) [a]
};
\addplot+[scatter, scatter src=explicit symbolic, only marks = {true}, scatter/classes = {{a={mark=*,red},b={mark=*,blue}}}] coordinates {
(2.4492935982947068e-18, -0.26) [b]
(-0.02372388652432624, 0.017599314699499942) [b]
(-0.11964922610498104, 0.3737100364667263) [b]
(-0.19948953853459253, 0.787850195426665) [b]
(-0.16405930325910698, 1.239519745641034) [b]
(-0.053272275009925624, 1.6993599685007101) [b]
(0.0972147056410751, 2.159369003258805) [b]
(0.1738525267036508, 2.641957840922275) [b]
(0.0942264873522882, 3.1279036831570197) [b]
(0.006416642694785807, 3.616097484945837) [b]
(0.003078243869734422, 4.014108132260943) [b]
};
\end{groupplot}

\end{tikzpicture}


%% file: active_query_process.tex
\newcommand{\asymcloud}[2][.1]{%
\begin{scope}[#2]
\pgftransformscale{#1}%
\pgfpathmoveto{\pgfpoint{261 pt}{115 pt}} 
  \pgfpathcurveto{\pgfqpoint{70 pt}{107 pt}}
                 {\pgfqpoint{137 pt}{291 pt}}
                 {\pgfqpoint{260 pt}{273 pt}} 
  \pgfpathcurveto{\pgfqpoint{78 pt}{382 pt}}
                 {\pgfqpoint{381 pt}{445 pt}}
                 {\pgfqpoint{412 pt}{410 pt}}
  \pgfpathcurveto{\pgfqpoint{577 pt}{587 pt}}
                 {\pgfqpoint{698 pt}{488 pt}}
                 {\pgfqpoint{685 pt}{366 pt}}
  \pgfpathcurveto{\pgfqpoint{840 pt}{192 pt}}
                 {\pgfqpoint{610 pt}{157 pt}}
                 {\pgfqpoint{610 pt}{157 pt}}
  \pgfpathcurveto{\pgfqpoint{531 pt}{39 pt}}
                 {\pgfqpoint{298 pt}{51 pt}}
                 {\pgfqpoint{261 pt}{115 pt}}
                 \pgfusepath{fill,stroke}         
\end{scope}}

\begin{tikzpicture}

 \node at (-2, 4) {\tikz[cm={-1,0,0,1,(0,0)}]\asymcloud[.08]{fill=gray!20,thick};}; 
 \node at (-1.88, 3.87) {$p(\textbf{w})$};
 \draw [->] (-1.0, 3.87) -- (0.6, 3.87);
 \node at (-0.15, 4.1) {sample};
 \node at (1.15, 3.7) 
 {
 	$\begin{aligned}
 		\begin{bmatrix}
 			\bf{w}_1 \\
 			\bf{w}_2 \\
 			\bf{w}_3 \\
 			\vdots \\
 			\bf{w}_M
 		\end{bmatrix}
 	\end{aligned}$
 };
 \draw [->] (1.7, 3.87) -- (3.3, 3.87);
 \node at (2.5, 4.1) {apply};
 \node at (2.5, 3.6) {heuristic};
 \node at (3.85, 3.87) 
 {
 	$\begin{aligned}
 		\begin{bmatrix}
 			\bf{w}_A \\
 			\bf{w}_B
 		\end{bmatrix}
 	\end{aligned}$
 };
 \draw [-] (4.4, 3.87) -- (4.8, 3.87);
 \draw [-] (4.8, 3.87) -- (4.8, 1.8);
 \draw [->] (4.8, 1.8) -- (3.1, 1.8);
 \node at (3.95, 2.1) {optimize};
 \node at (2.6, 1.8) 
 {
 	$\begin{aligned}
 		\begin{bmatrix}
 			{\tau}_A \\
 			{\tau}_B
 		\end{bmatrix}
 	\end{aligned}$
 };
 \node at (1.25, 2.1) {query};
 \node at (1.25, 1.6) {human};
 \draw [->] (2.1, 1.8) -- (0.4, 1.8);
 \node[circle,fill,minimum size=3mm] at (-0.2,2.5) (head) {};
 \node[rounded corners=2pt,minimum height=1.0cm,minimum width=0.3cm,fill,below = 1pt of head] (body) {};
 \draw[line width=1mm,round cap-round cap] ([shift={(2pt,-1pt)}]body.north east) --++(-90:5mm);
 \draw[line width=0.75mm,round cap-round cap] ([shift={(-2pt,-1pt)}]body.north west)--++(-90:5mm);
 \draw[thick,white,-round cap] (body.south) --++(90:5.5mm);
 \node at (-2.1, 2.1) {update $p(\textbf{w})$};
 \draw [->] (-3.4, 3.87) -- (-3.0, 3.87);
 \draw [-] (-3.4, 1.8) -- (-3.4, 3.87);
 \draw [-] (-0.8, 1.8) -- (-3.4, 1.8);
\end{tikzpicture}

%% file: compute_prob.tex
\begin{tikzpicture}[thick]

  \node at (-1.6,-0.5) {$\tau_A^{(n)}$};
  \draw [->] (-1.3, -0.5) -- (-0.7,0.0);
  \draw [->] (-1.3, -0.5) -- (-0.7,-0.9);
  
  \draw (-0.6,-0.3)  rectangle (0.2, 0.3);
  \node at (-0.2,0.0) {$\phi_{hc}$};
  
  \draw (-0.6,-1.2)  rectangle (0.2, -0.6);  
  \node at (-0.2,-0.9) {$\phi_{nn}$};
  
  \draw [-] (0.3, -0.9) -- (0.8,-0.5);
  \draw [-] (0.3, 0.0) -- (0.8,-0.5);
  \draw [->] (0.8, -0.5) -- (1.1,-0.5);
  \node at (1.4,-0.5) {$\Phi_{A}$};
  
  \draw [->] (1.7,-0.5) -- (2.2, -0.5);
  \draw (2.6, -0.5) circle (7pt);
  \node at (2.6, -0.5) {$\times$};
  
  \draw [->] (2.6, -1.15) -- (2.6, -0.85);
  \node at (2.6, -1.3) {\textbf{w}};

  \draw [-] (3.0,-0.5) -- (3.6, -1.3);  
  
  \node at (3.6, -0.75) {\footnotesize $R_A$};
  
  \node at (-1.6,-2.1) {$\tau_B^{(n)}$};
  \draw [->] (-1.3, -2.1) -- (-0.7,-1.6);
  \draw [->] (-1.3, -2.1) -- (-0.7,-2.5);
  
  \draw (-0.6,-2.3)  rectangle (0.2, -2.9);
  \node at (-0.2,-2.6) {$\phi_{nn}$};
  
  \draw (-0.6,-1.4)  rectangle (0.2, -2.0);  
  \node at (-0.2,-1.7) {$\phi_{hc}$};
  
  \draw [-] (0.3, -1.7) -- (0.8,-2.1);
  \draw [-] (0.3, -2.6) -- (0.8,-2.1);
  \draw [->] (0.8, -2.1) -- (1.1,-2.1);
  \node at (1.4,-2.1) {$\Phi_{B}$};
  
  \draw [->] (1.7,-2.1) -- (2.2, -2.1);
  \draw (2.6, -2.1) circle (7pt);
  \node at (2.6, -2.1) {$\times$};
  
  \draw [->] (2.6, -1.45) -- (2.6, -1.75);

  \draw [-] (3.0,-2.1) -- (3.6, -1.3);    
  
  \node at (3.6, -1.85) {\footnotesize $R_B$};
  
  
 \draw [->] (3.6,-1.3) -- (3.8, -1.3); 
 \node at (5.15,-1.3) {\footnotesize $P_A^{(n)} = \sigma(R_A - R_B)$};
  
\end{tikzpicture}

%% file: accuracy_chart.tex
\begin{tikzpicture}[]
\begin{axis}[height = {5cm}, ylabel = {Test accuracy}, xlabel = {User number}, ybar=0pt, bar width=1pt, xtick=data, symbolic x coords={1, 2, 3, 4, 5, 6, 7, 8, 9, 10, 11, 12, 13, 14, 15},, ymin = {0}, width = {8cm},  
legend style={at={(1,1.3), font=\footnotesize}}]
\addplot+ [bar width=4pt]coordinates {
(1, 0.5867)
(2, 0.7067)
(3, 0.7467)
(4, 0.6267)
(5, 0.6133)
(6, 0.5733)
(7, 0.6267)
(8, 0.5600)
(9, 0.4933)
(10, 0.5733)
(11, 0.5600)
(12, 0.6800)
(13, 0.6667)
(14, 0.5857)
(15, 0.5733)
};
\addlegendentry{hand-coded}
\addplot+ [
bar width=4pt, error bars/.cd, 
x dir=both, x explicit, y dir=both, y explicit]
table [
x error plus=ex+, x error minus=ex-, y error plus=ey+, y error minus=ey-
] {
x y ex+ ex- ey+ ey-
1 0.672 0.0 0.0 0.02022 0.02022
2 0.784 0.0 0.0 0.01738 0.01738
3 0.667 0.0 0.0 0.01333 0.01333
4 0.651 0.0 0.0 0.06422 0.06422
5 0.672 0.0 0.0 0.0073 0.0073
6 0.664 0.0 0.0 0.01115 0.01115
7 0.733 0.0 0.0 0.01333 0.01333
8 0.667 0.0 0.0 0.0163 0.0163
9 0.693 0.0 0.0 0.0133 0.0133
10 0.792 0.0 0.0 0.0152 0.0152
11 0.7547 0.0 0.0 0.0223 0.0223
12 0.784 0.0 0.0 0.03451 0.03451
13 0.696 0.0 0.0 0.01116 0.01116
14 0.7707 0.0 0.0 0.0112 0.0112
15 0.7067 0.0 0.0 0.03528 0.03528
};
\addlegendentry{mixed}
\end{axis}

\end{tikzpicture}

%% file: heading_both.tex
\begin{tikzpicture}[]
\begin{axis}[height = {5.5cm}, ylabel = {Feature Value}, xmin = {0}, xmax = {360}, ymax = {1}, xlabel = {$\theta_r$}, ymin = {-1}, width = {6.5cm}, legend style={font=\footnotesize}]\addplot+ [mark = {none}, style = {thick, pastelBlue}]coordinates {
(0.0, -0.5933285)
(2.0, -0.57793504)
(4.0, -0.5622438)
(6.0, -0.54617035)
(8.0, -0.5306935)
(10.0, -0.51503855)
(12.0, -0.49485776)
(14.0, -0.47306913)
(16.0, -0.45248976)
(18.0, -0.4335231)
(20.0, -0.41416448)
(22.0, -0.39581755)
(24.0, -0.3776116)
(26.0, -0.3550751)
(28.0, -0.3317003)
(30.0, -0.30686718)
(32.0, -0.28051832)
(34.0, -0.2537397)
(36.0, -0.22656636)
(38.0, -0.19903578)
(40.0, -0.17118737)
(42.0, -0.14325455)
(44.0, -0.11870083)
(46.0, -0.094001)
(48.0, -0.06918495)
(50.0, -0.04251597)
(52.0, -0.01424436)
(54.0, 0.014050244)
(56.0, 0.042322192)
(58.0, 0.07052656)
(60.0, 0.09813317)
(62.0, 0.12514749)
(64.0, 0.15037021)
(66.0, 0.1709202)
(68.0, 0.19132248)
(70.0, 0.21156079)
(72.0, 0.2316194)
(74.0, 0.25148284)
(76.0, 0.27113676)
(78.0, 0.2905672)
(80.0, 0.3097609)
(82.0, 0.32870528)
(84.0, 0.34738898)
(86.0, 0.36085537)
(88.0, 0.36358508)
(90.0, 0.3429794)
(92.0, 0.31803)
(94.0, 0.28902084)
(96.0, 0.25947112)
(98.0, 0.22962102)
(100.0, 0.19936676)
(102.0, 0.16872753)
(104.0, 0.13775896)
(106.0, 0.10559177)
(108.0, 0.068459146)
(110.0, 0.029428445)
(112.0, -0.009692369)
(114.0, -0.048783187)
(116.0, -0.088546)
(118.0, -0.129116)
(120.0, -0.16925819)
(122.0, -0.20852663)
(124.0, -0.24668407)
(126.0, -0.28409177)
(128.0, -0.32065442)
(130.0, -0.35628548)
(132.0, -0.39090976)
(134.0, -0.424462)
(136.0, -0.45688748)
(138.0, -0.4881427)
(140.0, -0.5181941)
(142.0, -0.5470181)
(144.0, -0.5746011)
(146.0, -0.6009377)
(148.0, -0.62603086)
(150.0, -0.64989156)
(152.0, -0.67253625)
(154.0, -0.6939879)
(156.0, -0.7142744)
(158.0, -0.7334277)
(160.0, -0.75148314)
(162.0, -0.76847905)
(164.0, -0.7844557)
(166.0, -0.7994547)
(168.0, -0.81351924)
(170.0, -0.8266923)
(172.0, -0.83901745)
(174.0, -0.85053784)
(176.0, -0.86129594)
(178.0, -0.8713335)
(180.0, -0.8806913)
(182.0, -0.8894086)
(184.0, -0.89752376)
(186.0, -0.90507334)
(188.0, -0.91209245)
(190.0, -0.91861486)
(192.0, -0.9246723)
(194.0, -0.9302953)
(196.0, -0.93551254)
(198.0, -0.9403514)
(200.0, -0.94483757)
(202.0, -0.9489952)
(204.0, -0.952847)
(206.0, -0.95641446)
(208.0, -0.9597176)
(210.0, -0.9627752)
(212.0, -0.96560466)
(214.0, -0.9682227)
(216.0, -0.9706443)
(218.0, -0.97288394)
(220.0, -0.9749549)
(222.0, -0.9768696)
(224.0, -0.9786394)
(226.0, -0.9802752)
(228.0, -0.9817869)
(230.0, -0.98318374)
(232.0, -0.9844742)
(234.0, -0.9856664)
(236.0, -0.98676765)
(238.0, -0.98778486)
(240.0, -0.9887243)
(242.0, -0.98959184)
(244.0, -0.9903929)
(246.0, -0.9911327)
(248.0, -0.9918157)
(250.0, -0.9924463)
(252.0, -0.9930285)
(254.0, -0.993566)
(256.0, -0.9940621)
(258.0, -0.99452007)
(260.0, -0.99494284)
(262.0, -0.9953331)
(264.0, -0.99569327)
(266.0, -0.9960257)
(268.0, -0.9963325)
(270.0, -0.99661565)
(272.0, -0.996877)
(274.0, -0.99711823)
(276.0, -0.9973408)
(278.0, -0.9975462)
(280.0, -0.9977358)
(282.0, -0.99791074)
(284.0, -0.99807215)
(286.0, -0.9982211)
(288.0, -0.9983586)
(290.0, -0.99848545)
(292.0, -0.9986025)
(294.0, -0.9987105)
(296.0, -0.9988102)
(298.0, -0.99890214)
(300.0, -0.998987)
(302.0, -0.99906534)
(304.0, -0.9991376)
(306.0, -0.9992043)
(308.0, -0.9992658)
(310.0, -0.99932253)
(312.0, -0.9993749)
(314.0, -0.99942327)
(316.0, -0.99946785)
(318.0, -0.99950904)
(320.0, -0.999547)
(322.0, -0.999582)
(324.0, -0.99961436)
(326.0, -0.99964416)
(328.0, -0.9996717)
(330.0, -0.9996971)
(332.0, -0.9997205)
(334.0, -0.999742)
(336.0, -0.9997618)
(338.0, -0.9997801)
(340.0, -0.999797)
(342.0, -0.9998126)
(344.0, -0.99982697)
(346.0, -0.99984026)
(348.0, -0.99985254)
(350.0, -0.99986386)
(352.0, -0.99987435)
(354.0, -0.999884)
(356.0, -0.9998929)
(358.0, -0.9999011)
(360.0, -0.99990875)
};
\addlegendentry{neural network}
\addplot+ [mark = {none}, style = {thick, pastelRed}]coordinates {
(0.0, 0.0)
(2.0, 0.03489949670250097)
(4.0, 0.0697564737441253)
(6.0, 0.10452846326765347)
(8.0, 0.13917310096006544)
(10.0, 0.17364817766693036)
(12.0, 0.20791169081775934)
(14.0, 0.24192189559966773)
(16.0, 0.27563735581699916)
(18.0, 0.30901699437494745)
(20.0, 0.3420201433256687)
(22.0, 0.374606593415912)
(24.0, 0.4067366430758002)
(26.0, 0.4383711467890774)
(28.0, 0.46947156278589075)
(30.0, 0.5)
(32.0, 0.5299192642332049)
(34.0, 0.5591929034707468)
(36.0, 0.5877852522924731)
(38.0, 0.6156614753256583)
(40.0, 0.6427876096865394)
(42.0, 0.6691306063588582)
(44.0, 0.6946583704589973)
(46.0, 0.7193398003386511)
(48.0, 0.7431448254773942)
(50.0, 0.766044443118978)
(52.0, 0.788010753606722)
(54.0, 0.8090169943749475)
(56.0, 0.8290375725550417)
(58.0, 0.848048096156426)
(60.0, 0.8660254037844386)
(62.0, 0.882947592858927)
(64.0, 0.898794046299167)
(66.0, 0.9135454576426009)
(68.0, 0.9271838545667874)
(70.0, 0.9396926207859084)
(72.0, 0.9510565162951535)
(74.0, 0.9612616959383189)
(76.0, 0.9702957262759965)
(78.0, 0.9781476007338057)
(80.0, 0.984807753012208)
(82.0, 0.9902680687415704)
(84.0, 0.9945218953682733)
(86.0, 0.9975640502598242)
(88.0, 0.9993908270190958)
(90.0, 1.0)
(92.0, 0.9993908270190958)
(94.0, 0.9975640502598242)
(96.0, 0.9945218953682733)
(98.0, 0.9902680687415704)
(100.0, 0.984807753012208)
(102.0, 0.9781476007338057)
(104.0, 0.9702957262759965)
(106.0, 0.9612616959383189)
(108.0, 0.9510565162951535)
(110.0, 0.9396926207859084)
(112.0, 0.9271838545667874)
(114.0, 0.9135454576426009)
(116.0, 0.898794046299167)
(118.0, 0.882947592858927)
(120.0, 0.8660254037844386)
(122.0, 0.848048096156426)
(124.0, 0.8290375725550417)
(126.0, 0.8090169943749475)
(128.0, 0.788010753606722)
(130.0, 0.766044443118978)
(132.0, 0.7431448254773942)
(134.0, 0.7193398003386511)
(136.0, 0.6946583704589973)
(138.0, 0.6691306063588582)
(140.0, 0.6427876096865394)
(142.0, 0.6156614753256583)
(144.0, 0.5877852522924731)
(146.0, 0.5591929034707468)
(148.0, 0.5299192642332049)
(150.0, 0.5)
(152.0, 0.46947156278589075)
(154.0, 0.4383711467890774)
(156.0, 0.4067366430758002)
(158.0, 0.374606593415912)
(160.0, 0.3420201433256687)
(162.0, 0.30901699437494745)
(164.0, 0.27563735581699916)
(166.0, 0.24192189559966773)
(168.0, 0.20791169081775934)
(170.0, 0.17364817766693036)
(172.0, 0.13917310096006544)
(174.0, 0.10452846326765347)
(176.0, 0.0697564737441253)
(178.0, 0.03489949670250097)
(180.0, 0.0)
(182.0, -0.03489949670250097)
(184.0, -0.0697564737441253)
(186.0, -0.10452846326765347)
(188.0, -0.13917310096006544)
(190.0, -0.17364817766693036)
(192.0, -0.20791169081775934)
(194.0, -0.24192189559966773)
(196.0, -0.27563735581699916)
(198.0, -0.30901699437494745)
(200.0, -0.3420201433256687)
(202.0, -0.374606593415912)
(204.0, -0.4067366430758002)
(206.0, -0.4383711467890774)
(208.0, -0.46947156278589075)
(210.0, -0.5)
(212.0, -0.5299192642332049)
(214.0, -0.5591929034707468)
(216.0, -0.5877852522924731)
(218.0, -0.6156614753256583)
(220.0, -0.6427876096865394)
(222.0, -0.6691306063588582)
(224.0, -0.6946583704589973)
(226.0, -0.7193398003386511)
(228.0, -0.7431448254773942)
(230.0, -0.766044443118978)
(232.0, -0.788010753606722)
(234.0, -0.8090169943749475)
(236.0, -0.8290375725550417)
(238.0, -0.848048096156426)
(240.0, -0.8660254037844386)
(242.0, -0.882947592858927)
(244.0, -0.898794046299167)
(246.0, -0.9135454576426009)
(248.0, -0.9271838545667874)
(250.0, -0.9396926207859084)
(252.0, -0.9510565162951535)
(254.0, -0.9612616959383189)
(256.0, -0.9702957262759965)
(258.0, -0.9781476007338057)
(260.0, -0.984807753012208)
(262.0, -0.9902680687415704)
(264.0, -0.9945218953682733)
(266.0, -0.9975640502598242)
(268.0, -0.9993908270190958)
(270.0, -1.0)
(272.0, -0.9993908270190958)
(274.0, -0.9975640502598242)
(276.0, -0.9945218953682733)
(278.0, -0.9902680687415704)
(280.0, -0.984807753012208)
(282.0, -0.9781476007338057)
(284.0, -0.9702957262759965)
(286.0, -0.9612616959383189)
(288.0, -0.9510565162951535)
(290.0, -0.9396926207859084)
(292.0, -0.9271838545667874)
(294.0, -0.9135454576426009)
(296.0, -0.898794046299167)
(298.0, -0.882947592858927)
(300.0, -0.8660254037844386)
(302.0, -0.848048096156426)
(304.0, -0.8290375725550417)
(306.0, -0.8090169943749475)
(308.0, -0.788010753606722)
(310.0, -0.766044443118978)
(312.0, -0.7431448254773942)
(314.0, -0.7193398003386511)
(316.0, -0.6946583704589973)
(318.0, -0.6691306063588582)
(320.0, -0.6427876096865394)
(322.0, -0.6156614753256583)
(324.0, -0.5877852522924731)
(326.0, -0.5591929034707468)
(328.0, -0.5299192642332049)
(330.0, -0.5)
(332.0, -0.46947156278589075)
(334.0, -0.4383711467890774)
(336.0, -0.4067366430758002)
(338.0, -0.374606593415912)
(340.0, -0.3420201433256687)
(342.0, -0.30901699437494745)
(344.0, -0.27563735581699916)
(346.0, -0.24192189559966773)
(348.0, -0.20791169081775934)
(350.0, -0.17364817766693036)
(352.0, -0.13917310096006544)
(354.0, -0.10452846326765347)
(356.0, -0.0697564737441253)
(358.0, -0.03489949670250097)
(360.0, 0.0)
};
\addlegendentry{hand-coded}
\addplot+ [mark = {none}, black,dashed]coordinates {
(90, -1)
(90, 1)
};
\end{axis}

\end{tikzpicture}

%% file: feature_viz_2d.tex
\begin{tikzpicture}[]
\begin{groupplot}[group style={horizontal sep = 0.2cm, vertical sep = 1.5cm, group size=5 by 1}]
\nextgroupplot [height = {7cm}, xmin = {-0.5}, xmax = {0.5}, axis equal image = {true}, ymax = {2.8}, ymin = {-1.8}, width = {3cm}, enlargelimits = false, axis on top, yticklabels={,,}, xticklabels={,,}]\addplot [point meta min=-0.22695684, point meta max=0.34060514] graphics [xmin=-0.5, xmax=0.5, ymin=-1.8, ymax=2.8] {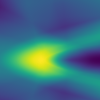};
\node[sedan top,body color=red,window color=black, minimum width=1.5cm,rotate=90.0,scale = 0.2] at (axis cs:0.17, 0.041) {};;
\addplot+ [mark = {none}, white,solid,thick]coordinates {
(-0.255, -1.8)
(-0.255, 2.8)
};
\addplot+ [mark = {none}, white,solid,thick]coordinates {
(0.255, -1.8)
(0.255, 2.8)
};
\addplot+ [mark = {none}, white,dashed,thick]coordinates {
(-0.085, -1.8)
(-0.085, 2.8)
};
\addplot+ [mark = {none}, white,dashed,thick]coordinates {
(0.085, -1.8)
(0.085, 2.8)
};
\nextgroupplot [height = {7cm}, xmin = {-0.5}, xmax = {0.5}, axis equal image = {true}, ymax = {2.8}, ymin = {-1.8}, width = {3cm}, enlargelimits = false, axis on top, yticklabels={,,}, xticklabels={,,}]\addplot [point meta min=-0.1959982, point meta max=0.34057143] graphics [xmin=-0.5, xmax=0.5, ymin=-1.8, ymax=2.8] {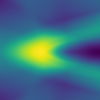};
\node[sedan top,body color=red,window color=black, minimum width=1.5cm,rotate=90.0,scale = 0.2] at (axis cs:0.17, 0.41) {};;
\addplot+ [mark = {none}, white,solid,thick]coordinates {
(-0.255, -1.8)
(-0.255, 2.8)
};
\addplot+ [mark = {none}, white,solid,thick]coordinates {
(0.255, -1.8)
(0.255, 2.8)
};
\addplot+ [mark = {none}, white,dashed,thick]coordinates {
(-0.085, -1.8)
(-0.085, 2.8)
};
\addplot+ [mark = {none}, white,dashed,thick]coordinates {
(0.085, -1.8)
(0.085, 2.8)
};
\nextgroupplot [height = {7cm}, xmin = {-0.5}, xmax = {0.5}, axis equal image = {true}, ymax = {2.8}, ymin = {-1.8}, width = {3cm}, enlargelimits = false, axis on top, yticklabels={,,}, xticklabels={,,}]\addplot [point meta min=-0.21508682, point meta max=0.32442644] graphics [xmin=-0.5, xmax=0.5, ymin=-1.8, ymax=2.8] {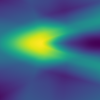};
\addplot+ [mark = {none}, white,solid,thick]coordinates {
(-0.255, -1.8)
(-0.255, 2.8)
};
\addplot+ [mark = {none}, white,solid,thick]coordinates {
(0.255, -1.8)
(0.255, 2.8)
};
\addplot+ [mark = {none}, white,dashed,thick]coordinates {
(-0.085, -1.8)
(-0.085, 2.8)
};
\addplot+ [mark = {none}, white,dashed,thick]coordinates {
(0.085, -1.8)
(0.085, 2.8)
};
\node[sedan top,body color=red,window color=black, minimum width=1.5cm,rotate=113.49126960036371,scale = 0.2] at (axis cs:0.09530392435697597, 0.8102525110288488) {};;
\nextgroupplot [height = {7cm}, xmin = {-0.5}, xmax = {0.5}, axis equal image = {true}, ymax = {2.8}, ymin = {-1.8}, width = {3cm}, enlargelimits = false, axis on top, yticklabels={,,}, xticklabels={,,}]\addplot [point meta min=-0.24934933, point meta max=0.30432513] graphics [xmin=-0.5, xmax=0.5, ymin=-1.8, ymax=2.8] {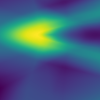};
\node[sedan top,body color=red,window color=black, minimum width=1.5cm,rotate=90.0,scale = 0.2] at (axis cs:0.0042648662665906506, 1.2071069757931783) {};;
\addplot+ [mark = {none}, white,solid,thick]coordinates {
(-0.255, -1.8)
(-0.255, 2.8)
};
\addplot+ [mark = {none}, white,solid,thick]coordinates {
(0.255, -1.8)
(0.255, 2.8)
};
\addplot+ [mark = {none}, white,dashed,thick]coordinates {
(-0.085, -1.8)
(-0.085, 2.8)
};
\addplot+ [mark = {none}, white,dashed,thick]coordinates {
(0.085, -1.8)
(0.085, 2.8)
};
\nextgroupplot [height = {7cm}, xmin = {-0.5}, xmax = {0.5}, axis equal image = {true}, ymax = {2.8}, ymin = {-1.8}, width = {3cm}, enlargelimits = false, axis on top, yticklabels={,,}, xticklabels={,,}]\addplot [point meta min=-0.27402392, point meta max=0.3042589] graphics [xmin=-0.5, xmax=0.5, ymin=-1.8, ymax=2.8] {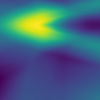};
\node[sedan top,body color=red,window color=black, minimum width=1.5cm,rotate=90.0,scale = 0.2] at (axis cs:0.004264866266590677, 1.6599944239254785) {};;
\addplot+ [mark = {none}, white,solid,thick]coordinates {
(-0.255, -1.8)
(-0.255, 2.8)
};
\addplot+ [mark = {none}, white,solid,thick]coordinates {
(0.255, -1.8)
(0.255, 2.8)
};
\addplot+ [mark = {none}, white,dashed,thick]coordinates {
(-0.085, -1.8)
(-0.085, 2.8)
};
\addplot+ [mark = {none}, white,dashed,thick]coordinates {
(0.085, -1.8)
(0.085, 2.8)
};
\end{groupplot}

\end{tikzpicture}

%% file: feature_viz_2d_time_gap.tex
\begin{tikzpicture}[]
\begin{groupplot}[group style={horizontal sep = 0.2cm, vertical sep = 1.5cm, group size=5 by 1}]
\nextgroupplot [height = {7cm}, xmin = {-0.5}, xmax = {0.5}, axis equal image = {true}, ymax = {2.8}, ymin = {-1.8}, width = {3cm}, enlargelimits = false, axis on top, yticklabels={,,}, xticklabels={,,}]\addplot [point meta min=-0.22695684, point meta max=0.34060514] graphics [xmin=-0.5, xmax=0.5, ymin=-1.8, ymax=2.8] {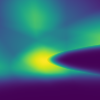};
\node[sedan top,body color=red,window color=black, minimum width=1.5cm,rotate=90.0,scale = 0.2] at (axis cs:0.17, 0.041) {};;
\addplot+ [mark = {none}, white,solid,thick]coordinates {
(-0.255, -1.8)
(-0.255, 2.8)
};
\addplot+ [mark = {none}, white,solid,thick]coordinates {
(0.255, -1.8)
(0.255, 2.8)
};
\addplot+ [mark = {none}, white,dashed,thick]coordinates {
(-0.085, -1.8)
(-0.085, 2.8)
};
\addplot+ [mark = {none}, white,dashed,thick]coordinates {
(0.085, -1.8)
(0.085, 2.8)
};
\nextgroupplot [height = {7cm}, xmin = {-0.5}, xmax = {0.5}, axis equal image = {true}, ymax = {2.8}, ymin = {-1.8}, width = {3cm}, enlargelimits = false, axis on top, yticklabels={,,}, xticklabels={,,}]\addplot [point meta min=-0.1959982, point meta max=0.34057143] graphics [xmin=-0.5, xmax=0.5, ymin=-1.8, ymax=2.8] {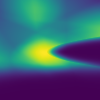};
\node[sedan top,body color=red,window color=black, minimum width=1.5cm,rotate=90.0,scale = 0.2] at (axis cs:0.17, 0.41) {};;
\addplot+ [mark = {none}, white,solid,thick]coordinates {
(-0.255, -1.8)
(-0.255, 2.8)
};
\addplot+ [mark = {none}, white,solid,thick]coordinates {
(0.255, -1.8)
(0.255, 2.8)
};
\addplot+ [mark = {none}, white,dashed,thick]coordinates {
(-0.085, -1.8)
(-0.085, 2.8)
};
\addplot+ [mark = {none}, white,dashed,thick]coordinates {
(0.085, -1.8)
(0.085, 2.8)
};
\nextgroupplot [height = {7cm}, xmin = {-0.5}, xmax = {0.5}, axis equal image = {true}, ymax = {2.8}, ymin = {-1.8}, width = {3cm}, enlargelimits = false, axis on top, yticklabels={,,}, xticklabels={,,}]\addplot [point meta min=-0.21508682, point meta max=0.32442644] graphics [xmin=-0.5, xmax=0.5, ymin=-1.8, ymax=2.8] {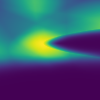};
\addplot+ [mark = {none}, white,solid,thick]coordinates {
(-0.255, -1.8)
(-0.255, 2.8)
};
\addplot+ [mark = {none}, white,solid,thick]coordinates {
(0.255, -1.8)
(0.255, 2.8)
};
\addplot+ [mark = {none}, white,dashed,thick]coordinates {
(-0.085, -1.8)
(-0.085, 2.8)
};
\addplot+ [mark = {none}, white,dashed,thick]coordinates {
(0.085, -1.8)
(0.085, 2.8)
};
\node[sedan top,body color=red,window color=black, minimum width=1.5cm,rotate=113.49126960036371,scale = 0.2] at (axis cs:0.09530392435697597, 0.8102525110288488) {};;
\nextgroupplot [height = {7cm}, xmin = {-0.5}, xmax = {0.5}, axis equal image = {true}, ymax = {2.8}, ymin = {-1.8}, width = {3cm}, enlargelimits = false, axis on top, yticklabels={,,}, xticklabels={,,}]\addplot [point meta min=-0.24934933, point meta max=0.30432513] graphics [xmin=-0.5, xmax=0.5, ymin=-1.8, ymax=2.8] {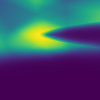};
\node[sedan top,body color=red,window color=black, minimum width=1.5cm,rotate=90.0,scale = 0.2] at (axis cs:0.0042648662665906506, 1.2071069757931783) {};;
\addplot+ [mark = {none}, white,solid,thick]coordinates {
(-0.255, -1.8)
(-0.255, 2.8)
};
\addplot+ [mark = {none}, white,solid,thick]coordinates {
(0.255, -1.8)
(0.255, 2.8)
};
\addplot+ [mark = {none}, white,dashed,thick]coordinates {
(-0.085, -1.8)
(-0.085, 2.8)
};
\addplot+ [mark = {none}, white,dashed,thick]coordinates {
(0.085, -1.8)
(0.085, 2.8)
};
\nextgroupplot [height = {7cm}, xmin = {-0.5}, xmax = {0.5}, axis equal image = {true}, ymax = {2.8}, ymin = {-1.8}, width = {3cm}, enlargelimits = false, axis on top, yticklabels={,,}, xticklabels={,,}]\addplot [point meta min=-0.27402392, point meta max=0.3042589] graphics [xmin=-0.5, xmax=0.5, ymin=-1.8, ymax=2.8] {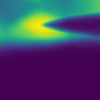};
\node[sedan top,body color=red,window color=black, minimum width=1.5cm,rotate=90.0,scale = 0.2] at (axis cs:0.004264866266590677, 1.6599944239254785) {};;
\addplot+ [mark = {none}, white,solid,thick]coordinates {
(-0.255, -1.8)
(-0.255, 2.8)
};
\addplot+ [mark = {none}, white,solid,thick]coordinates {
(0.255, -1.8)
(0.255, 2.8)
};
\addplot+ [mark = {none}, white,dashed,thick]coordinates {
(-0.085, -1.8)
(-0.085, 2.8)
};
\addplot+ [mark = {none}, white,dashed,thick]coordinates {
(0.085, -1.8)
(0.085, 2.8)
};
\end{groupplot}

\end{tikzpicture}